\documentclass[10pt,twocolumn,letterpaper]{article}

\usepackage{cvpr}

\usepackage{url}

\usepackage{graphicx}
\usepackage{booktabs}

\usepackage[]{amsmath}                  
\usepackage[]{mathtools}                
\usepackage[]{amssymb,amsfonts}
\usepackage{graphicx}
\usepackage{bm}

\usepackage[]{xcolor}
\usepackage[]{eucal}

\usepackage[]{xparse}
\usepackage[]{xstring}
\usepackage{etoolbox}
\usepackage[]{xspace}
\usepackage[]{comment}
\usepackage{float}

\usepackage{nicefrac}
\usepackage{amsmath}

\usepackage{multirow}
\usepackage{url}
\usepackage{tcolorbox}
\usepackage{graphicx}

\newcommand{\gray}[1]{\textcolor{gray}{#1}}

\newcommand{\R}{\mathbb{R}}

\newcommand{\M}{\mathcal{M}}
\newcommand{\E}{\mathbb{E}}
\newcommand{\T}{\mathcal{T}}

\newcommand{\image}{\mathbf{I}}
\newcommand{\depth}{\mathbf{D}}
\newcommand{\cam}{\mathbf{g}}
\newcommand{\rot}{\mathbf{r}}
\newcommand{\rotmatrix}{\mathbf{R}}
\newcommand{\intr}{\mathbf{K}}
\newcommand{\trans}{\mathbf{t}}
\newcommand{\fov}{\boldsymbol{\theta}}

\newcommand{\point}{\mathbf{P}}
\newcommand{\feat}{\mathbf{f}}
\newcommand{\weight}{\mathbf{W}}
\newcommand{\anyinput}{\boldsymbol{\xi}}

\definecolor{cvprblue}{rgb}{0.21,0.49,0.74}
\usepackage[pagebackref,breaklinks,colorlinks,allcolors=cvprblue]{hyperref}

\usepackage[table]{xcolor}

\definecolor{rkbest}{rgb}{0.84,0.93,0.86}
\definecolor{rksecond}{rgb}{0.96,0.92,0.82}
\definecolor{rkgray}{gray}{0.87}

\usepackage{colortbl}
\usepackage{pgfmath}

\definecolor{BestColor}{rgb}{0.72,0.84,0.78}
\definecolor{MidColor}{rgb}{1.00,1.00,0.93}
\definecolor{WorstColor}{rgb}{0.94,0.75,0.77}
\definecolor{NAColor}{gray}{0.87}

\newcommand{\RankCell}[3]{%
  \begingroup
  \pgfmathsetmacro{\tRaw}{((#1)-1)/max((#2)-1,1)}%
  \pgfmathsetmacro{\t}{max(0,min(1,\tRaw))}%
  \ifdim \t pt < 0.5pt
    \pgfmathsetmacro{\u}{2*\t}%
    \pgfmathsetmacro{\R}{0.72 + \u*(1.00-0.72)}%
    \pgfmathsetmacro{\G}{0.84 + \u*(1.00-0.84)}%
    \pgfmathsetmacro{\B}{0.78 + \u*(0.93-0.78)}%
  \else
    \pgfmathsetmacro{\u}{2*(\t-0.5)}%
    \pgfmathsetmacro{\R}{1.00 + \u*(0.94-1.00)}%
    \pgfmathsetmacro{\G}{1.00 + \u*(0.75-1.00)}%
    \pgfmathsetmacro{\B}{0.93 + \u*(0.77-0.93)}%
  \fi
  \edef\ApplyCellColor{\noexpand\cellcolor[rgb]{\R,\G,\B}}%
  \ApplyCellColor #3%
  \endgroup
}

\newcommand{\AvgRankCell}[3]{%
  \begingroup
  \pgfmathsetmacro{\tRaw}{((#1)-(#2))/max((#3)-(#2),0.0001)}%
  \pgfmathsetmacro{\t}{max(0,min(1,\tRaw))}%
  \ifdim \t pt < 0.5pt
    \pgfmathsetmacro{\u}{2*\t}%
    \pgfmathsetmacro{\R}{0.72 + \u*(1.00-0.72)}%
    \pgfmathsetmacro{\G}{0.84 + \u*(1.00-0.84)}%
    \pgfmathsetmacro{\B}{0.78 + \u*(0.93-0.78)}%
  \else
    \pgfmathsetmacro{\u}{2*(\t-0.5)}%
    \pgfmathsetmacro{\R}{1.00 + \u*(0.94-1.00)}%
    \pgfmathsetmacro{\G}{1.00 + \u*(0.75-1.00)}%
    \pgfmathsetmacro{\B}{0.93 + \u*(0.77-0.93)}%
  \fi
  \edef\ApplyAvgColor{\noexpand\cellcolor[rgb]{\R,\G,\B}}%
  \ApplyAvgColor #1%
  \endgroup
}

\def\paperID{}
\def\confName{CVPR}
\def\confYear{2026}

\title{Unlocking the Power of Critical Factors for 3D Visual Geometry Estimation}

\author{Guangkai Xu$ ^{1}$\thanks{
Both authors contributed equally. $^\dag$ Corresponding author.
},
    Hua Geng$ ^{1*}$,
    Huanyi Zheng$ ^{1}$,
    Songyi Yin$ ^{1}$,
    Yanlong Sun$ ^{2}$,
    Hao Chen$ ^{1}$,
    Chunhua Shen$ ^{1,3\dag}$
\\[0.2cm]
$^1$ State Key Lab of CAD \& CG, Zhejiang University ~~
$ ^2$ Tsinghua University ~~
$ ^3$ Ant Group
}

\begin{document}
\maketitle

\begin{abstract}
	Feed-forward visual geometry estimation has recently made rapid progress. However, an important gap remains: multi-frame models usually produce better cross-frame consistency, yet they often underperform strong per-frame methods on single-frame accuracy.
	This observation motivates our systematic investigation into the critical factors driving model performance through rigorous ablation studies, which reveals several key insights:
	1) Scaling up data diversity and quality unlocks further performance gains even in state-of-the-art visual geometry estimation methods; 2) Commonly adopted confidence-aware loss and gradient-based loss mechanisms may unintentionally hinder performance; 3) Joint supervision through both per-sequence and per-frame alignment improves results, while local region alignment surprisingly degrades performance.
	Furthermore, we introduce two enhancements to integrate the advantages of optimization-based methods and high-resolution inputs: a consistency loss function that enforces alignment between depth maps, camera parameters, and point maps, and an efficient architectural design that leverages high-resolution information. 
	We integrate these designs into \textbf{CARVE}, a resolution-enhanced model for feed-forward visual geometry estimation. Experiments on point cloud reconstruction, video depth estimation, and camera pose/intrinsic estimation show that CARVE achieves strong and robust performance across diverse benchmarks. GitHub: \url{https://github.com/aim-uofa/CARVE}.

\end{abstract}

\section{Introduction}

Recovering accurate and consistent 3D attributes from monocular video, including 3D point clouds, camera parameters, and depth maps, remains a long-standing challenge with broad applications, such as autonomous driving \cite{chen20203d, zhu20123d}, virtual and augmented reality \cite{mahmood2020bim,placitelli2011low}, robotic navigation \cite{wang2021navigation, liu2015robotic}, and medical imaging \cite{starly2005three, cheng2021depth}.

Existing approaches can be roughly categorized into two groups: optimization‑based approaches and learning‑based approaches. Optimization‑based techniques \cite{schonberger2016structure, cui2017hsfm, pan2024global,moulon2013global} generally rely on robust feature matching to estimate 3D attributes by minimizing reprojection errors. These methods often produce sparse or semi-dense reconstructions due to the heavy dependence on reliable correspondences.

In contrast, learning-based approaches directly regress 3D attributes through end-to-end neural networks trained on large-scale labeled datasets, and can be broadly categorized into per-frame methods \cite{DAv2, Metric3Dv2, DepthPro, MoGe} and multi-frame methods \cite{VGGT, wang2025pi, depthanything3, fast3r, DUSt3R, MASt3R, spann3r, cut3r}.  
Despite having access to cross-frame information, multi-frame methods do not consistently outperform per-frame methods. In practice, multi-frame methods mainly improve temporal consistency, while per-frame approaches often achieve higher accuracy on individual frames. This advantage is often attributed to carefully designed training objectives, high-resolution inputs, and well-structured training curriculum. 

Motivated by these observations, we systematically investigate these key factors on a representative method VGGT \cite{VGGT} through extensive ablation studies, resulting in several critical insights:
1) Despite extensive pre-training on large datasets, scaling up data diversity and quality
unlocks further performance gain.
2) The commonly adopted spatial gradient loss and confidence-aware weighting strategy can unexpectedly degrade the model performance. Conversely, employing fixed weighting inversely proportional to ground-truth depth consistently improves performance;
3) The sequence-level and frame-level alignment strategy of training objectives improves overall performance, whereas the local region alignment unexpectedly brings a performance reduction.

Alongside these insights, we explore two complementary enhancements to integrate the advantage of optimization-based methods and leverage high-resolution inputs.
First, motivated by the geometric constraints of optimization-based techniques, we introduce a consistency loss enforcing strict consistency among estimated camera parameters, depth maps, and 3D point clouds. Second, rather than directly feeding high-resolution inputs, we propose to extract high-resolution and low-resolution ViT features, and fuse them via cross-attention equipped with zero-initialized gating parameters, which can preserve the pre-training knowledge.

Combining these insights and improvements, we scale up the training and propose \textbf{CARVE}, an accurate and resolution-enhanced visual geometry estimation model. The rigorous ablation study paves the way for better performance of video depth estimation, camera pose and intrinsics estimation, and 3D point cloud estimation on evaluation datasets including KITTI \cite{geiger2013vision}, 7-Scenes \cite{shotton2013scene}, TUM \cite{sturm2012benchmark}, HO3D \cite{hampali2020honnotate}, ETH3D \cite{schops2017multi}, HAMMER \cite{jung2023importance}, and Bonn \cite{palazzolo2019iros}. Our main contributions are summarized as follows:

\begin{itemize}

	\item Rigorous ablation experiments are conducted on a representative visual geometry method to explore how training objectives and data influence the performance.

	\item We propose a consistency loss to enforce geometric coherence among predicted camera parameters, depth maps, and 3D point clouds, thereby integrating intrinsic perspective projection constraints during training.

	\item We develop an efficient and effective feature fusion mechanism that integrates high-resolution features into low-resolution features via cross-attention, enabling accurate estimation with less computational burden.

	\item Integrating these insights and improvements, we propose the \textbf{CARVE} framework, achieving strong overall performance across diverse benchmark datasets, including point cloud reconstruction, video depth estimation, and camera pose/intrinsic estimation.

\end{itemize}

\section{Related Work}
\label{sec: related_work}

\subsection{Optimization-based Reconstruction}

Traditional Structure-from-Motion (SfM) \cite{moulon2013global, schonberger2016structure, cui2017hsfm, pan2024global}, Multi-View Stereo (MVS) \cite{schonberger2016pixelwise,goesele2007multi,furukawa2009accurate}, and visual SLAM methods \cite{mur2015orb, engel2014lsd, engel2017direct}, rely heavily on a multi-stage pipeline of feature extraction, matching, and optimization. They jointly estimate camera poses and per-pixel 3D geometry by minimizing the reprojection error. However, the geometric fidelity achieved by these optimization techniques is critically dependent on the accuracy and robustness of the initial feature correspondences. To enhance robustness, subsequent works \cite{DROID-SLAM, MegaSaM, yao2018mvsnet, huang2018deepmvs, ding2022transmvsnet} integrate learned features or correspondences into the optimization framework to assist or replace traditional modules.

\subsection{Per-frame Reconstruction}

One approach to 3D visual geometry estimation from monocular video is to first estimate per-frame geometry and then enforce multi-view consistency. Numerous studies have investigated monocular depth estimation, demonstrating notable progress and improved accuracy \cite{DA, DAv2, Metric3D, Metric3Dv2, DepthPro, LeReS, Marigold, xumatters, he2025diffcalib, fu2024geowizard}. Several of them \cite{Metric3D, DepthPro} can produce 3D reconstruction from a single image. The MoGe series \cite{MoGe,moge2} directly estimates dense affine-invariant point clouds. To ensure consistency between frames, these methods still rely on alignment with matching information \cite{sun2021loftr, sarlin2020superglue} or consistency optimization \cite{xu2023frozenrecon}.

\subsection{Multi-frame Reconstruction}

Moving beyond explicit matching cost volume, feed-forward geometry methods have emerged after matching-based stereo approaches \cite{PSMNet, liu2022local, LEAStereo}. DUSt3R \cite{DUSt3R} and its follow-ups \cite{MASt3R, Pow3R, FLARE} directly regress point clouds from input images using neural networks. Extending this paradigm to longer temporal contexts, Spann3R \cite{spann3r} and CUT3R \cite{cut3r} model multi-frame geometry with implicit scene representations. More recently, Fast3R \cite{fast3r}, VGGT \cite{VGGT}, and Pi3 \cite{wang2025pi} adopt feed-forward architectures to estimate multi-view geometry while reducing reliance on explicit correspondence matching and iterative optimization. In particular, Pi3 removes the need for a fixed reference view through a permutation-equivariant design, while Depth Anything 3 \cite{depthanything3} further generalizes visual geometry estimation with a simple transformer backbone and a unified depth-ray prediction target.

\section{Method}
\label{sec: method}

In this section, we first introduce the multi-frame baseline and then systematically investigate the gap between representative multi-frame and per-frame methods.

\subsection{Preliminaries}

\noindent\textbf{Problem Definition.} The representative visual geometry estimation method \cite{VGGT} takes a set of images $\image \in \R^{T \times H \times W \times 3}$ as input, and produces depth maps $\hat{\depth} \in \R^{T \times H \times W}$, world coordinates point maps $\hat{\point} \in \R^{T \times H \times W \times 3}$, and camera parameters $\hat{\cam} \in \R^{T \times 9}$, which are composed of quaternion $\hat{\rot} \in \R^{T \times 4}$, translation vectors $\hat{\trans} \in \R^{T \times 3}$, and field of view angles $\hat{\fov} \in \R^{T \times 2}$. The point tracking task is not discussed here.

\noindent\textbf{Network Architecture.} The network patchifies the input images $\image$ into tokens $\hat{\feat}_{\text{img}} \in \R^{T \times P \times C}$ with DINOv2 \cite{dinov2} encoder and then passes them along with learnable camera tokens $\feat_{\text{cam\_init}}$ into transformer blocks and decoders.
\begin{equation}
	\label{eq: model_vggt}
	\begin{gathered}
		\hat{\feat}_{\text{img}} = \mathrm{Encoder}(\image), ~(\hat{\feat}_{\text{geo}}, \hat{\feat}_{\text{cam}}) = \mathrm{Transformer}(\hat{\feat}_{\text{img}}, \feat_{\text{cam\_init}}), \\
		\hat{\depth} = \mathrm{Head}_{\text{depth}}(\hat{\feat}_{\text{geo}}),~~ \hat{\point} = \mathrm{Head}_{\text{point}}(\hat{\feat}_{\text{geo}})  \\
		\hat{\cam} = [\hat{\trans}, \hat{\rot}, \hat{\fov}] = \mathrm{Head}_{\text{cam}}(\hat{\feat}_{\text{cam}}).
	\end{gathered}
\end{equation}

\noindent\textbf{Training Objectives.} The training losses contain three types of functions. For regression loss, they filter out invalid regions and simply supervise the valid regions:
\begin{equation}
	\label{eq: l1_loss}
	\mathcal{L}_\text{reg}(\hat{\anyinput}, \anyinput, \weight) = \E_{p \in \M} \left\| \weight_{p} \cdot (\hat{\anyinput}_{p} - \anyinput_{p} ) \right\| ,
\end{equation}
where $\hat{\anyinput}$ and $\anyinput$ are the prediction and the ground truth of either depth maps or point maps. $\M$ represents the valid region, and $\weight$ means the weight map. For spatial gradient loss, it supervises the difference between nearby pixels:
\begin{equation}
	\label{eq: sg_loss}
	\mathcal{L}_\text{sg} (\hat{\anyinput}, \anyinput, \weight) = \E_{p \in \M} \left\| \weight_{p} \cdot (\nabla_p \hat{\anyinput}_{p} - \nabla_p \anyinput_{p})  \right\|,
\end{equation}
where $\nabla_p$ represents the difference between nearby pixels of the spatial x and y axes. Another confidence loss is adopted to supervise the learnable confidence map:
\begin{equation}
	\label{eq: conf_loss}
	\mathcal{L}_\text{conf} (\weight) = \E_{p \in \M} \left| - \alpha \log \weight_{p}  \right|.
\end{equation}

The overall training losses consist of three components:
\begin{equation}
	\begin{gathered}
		\label{eq: vggt_loss}
		\mathcal{L} = \mathcal{L}_\text{cam} + \mathcal{L}_\text{depth} + \mathcal{L}_\text{point},
		\mathcal{L}_\text{cam} = \E_t \left\| \hat{\cam}_t - \cam_t \right\|, \\
		\mathcal{L}_\text{depth} = \mathcal{L}_\text{reg}(\hat{\depth}, \depth, \boldsymbol{\Sigma}^{\text{d}}) + \mathcal{L}_\text{sg}(\hat{\depth}, \depth, \boldsymbol{\Sigma}^{\text{d}}) + \mathcal{L}_\text{conf}(\boldsymbol{\Sigma}^{\text{d}}), \\
		\mathcal{L}_\text{point} = \mathcal{L}_\text{reg}(\hat{\point}, \point, \boldsymbol{\Sigma}^{\text{p}}) + \mathcal{L}_\text{sg}(\hat{\point}, \point, \boldsymbol{\Sigma}^{\text{p}}) + \mathcal{L}_\text{conf}(\boldsymbol{\Sigma}^{\text{p}})
	\end{gathered}
\end{equation}
where the confidence maps $\boldsymbol{\Sigma}^{\text{d}}$ and $\boldsymbol{\Sigma}^{\text{p}}$ are learned automatically, serving as adaptive weights for the loss functions.

\subsection{Effectiveness of Training Components}

Compared with per-frame methods \cite{DAv2, Metric3Dv2, DepthPro, MoGe}, multi-frame visual geometry estimation methods \cite{VGGT,DUSt3R,MASt3R,spann3r, fast3r, cut3r} achieve better multi-frame consistency but lower per-frame accuracy. We conduct extensive ablations to investigate the reasons.

\noindent\textbf{Experimental Details.} By default, we initialize the model with VGGT \cite{VGGT} pretrained weights, freeze the ViT feature extractor, and train the remaining components. The predicted point cloud, depth map, and camera translation are aligned to ground truth via a per-sequence scale factor before loss computation. Training uses a dynamic batch size (up to 24 frames) for 30K iterations, and evaluation is conducted on uniformly sampled keyframes with up to 200 frames per video (see supplementary for details). For data ablation, compared to the original VGGT, only the training data is varied; for loss ablation, we fix ``Data3'' and evaluate different loss terms; and for resolution ablation, we use ``Data3'' with ``Our Loss''.

\noindent\textbf{Training Data.}
We progressively expand the training data from "Data1" to "Data3," with the composition summarized in Table \ref{tab: ablation_components_data}. Specifically, "Data1" consists solely of high-quality datasets, "Data2" introduces greater data diversity while maintaining quality, and "Data3" further incorporates noisy datasets. As shown in Table \ref{tab: ablation_study}, performance improves consistently with data scaling, suggesting that current visual geometry estimation models can still benefit from larger and more diverse training data.

\begin{tcolorbox}[
		colframe=black,
		arc=4pt,
		boxsep=1pt,
	]
	\paragraph{\textbf{\textit{Insight} 1.}}
	Scaling up data diversity and quality unlocks further performance gains in SOTA visual geometry estimation.
\end{tcolorbox}

\begin{table*}[th]
	\vspace{-1em}
	\centering
	\small
	\caption{Ablation study for training data, training loss, and high-resolution input. For training loss, we explore the effectiveness of the original VGGT losses ($\mathcal{L}_\text{sg}$, $\mathcal{L}_\text{conf}$), several losses adopted in state-of-the-art methods ($\mathcal{L}_\text{tg}$, $\mathcal{L}_\text{F}$, $\mathcal{L}_\text{S}$), and our proposed consistency loss ($\mathcal{L}_\text{consis}$). ``Rank'' represents the average rank value across all metrics. Rows in \gray{gray} denote training with up to 12 frames (vs. the usual 24) and evaluation with up to 100 frames (vs. the usual 200) due to GPU memory constraints.}
	\vspace{-1em}
	\renewcommand{\arraystretch}{0.9}
	\resizebox{\linewidth}{!}{
		\begin{tabular}{r |ccc| ccc| ccc | ccc| c}
			\toprule
			\multirow{3}{*}{\textbf{Method}} & \multicolumn{3}{c|}{\textbf{7-Scenes}} & \multicolumn{3}{c|}{\textbf{Bonn}} & \multicolumn{3}{c|}{\textbf{KITTI}} & \multicolumn{3}{c|}{\textbf{TUM}} & \multirow{3}{*}{\textbf{Rank↓}} \\
			\cmidrule(lr){2-4} \cmidrule(lr){5-7} \cmidrule(lr){8-10} \cmidrule(lr){11-13}
 & Recon & Pose & Depth & Recon & Pose & Depth & Recon & Pose & Depth & Recon & Pose & Depth & \\
 & C-L1↓ & ATE↓ & Rel↓ & C-L1↓ & ATE↓ & Rel↓ & C-L1↓ & ATE↓ & Rel↓ & C-L1↓ & ATE↓ & Rel↓ & \\
			\midrule
			VGGT baseline & 0.049 & 0.073 & 0.069 & 0.057 & 0.075 & 0.054 & 0.296 & 1.113 & 0.094 & 0.051 & 0.047 & 0.062 & -  \\

			\midrule
			Data1 & \RankCell{3}{3}{0.056} & \RankCell{3}{3}{0.079} & \RankCell{3}{3}{0.070} & \RankCell{2}{3}{0.051} & \RankCell{2}{3}{0.064} & \RankCell{2}{3}{0.049} & \RankCell{3}{3}{0.281} & \RankCell{3}{3}{1.411} & \RankCell{3}{3}{0.085} & \RankCell{2}{3}{0.040} & \RankCell{2}{3}{0.090} & \RankCell{2}{3}{0.049} & \AvgRankCell{2.50}{1.00}{2.50} \\
			Data2 & \RankCell{2}{3}{0.052} & \RankCell{2}{3}{0.078} & \RankCell{2}{3}{0.069} & \RankCell{2}{3}{0.051} & \RankCell{3}{3}{0.071} & \RankCell{3}{3}{0.052} & \RankCell{2}{3}{0.277} & \RankCell{2}{3}{1.267} & \RankCell{2}{3}{0.083} & \RankCell{2}{3}{0.040} & \RankCell{2}{3}{0.090} & \RankCell{3}{3}{0.052} & \AvgRankCell{2.25}{1.00}{2.50} \\
			(Our Data) Data3 & \RankCell{1}{3}{0.049} & \RankCell{1}{3}{0.065} & \RankCell{1}{3}{0.065} & \RankCell{1}{3}{0.048} & \RankCell{1}{3}{0.055} & \RankCell{1}{3}{0.046} & \RankCell{1}{3}{0.263} & \RankCell{1}{3}{0.937} & \RankCell{1}{3}{0.082} & \RankCell{1}{3}{0.038} & \RankCell{1}{3}{0.050} & \RankCell{1}{3}{0.042} & \AvgRankCell{1.00}{1.00}{2.50} \\
			\midrule
			(VGGT Loss) $\mathcal{L}_\text{reg}$ + $\mathcal{L}_\text{conf}$ + $\mathcal{L}_\text{sg}$ & \RankCell{2}{3}{0.049} & \RankCell{2}{3}{0.065} & \RankCell{2}{3}{0.065} & \RankCell{3}{3}{0.048} & \RankCell{3}{3}{0.055} & \RankCell{1}{3}{0.046} & \RankCell{2}{3}{0.263} & \RankCell{2}{3}{0.937} & \RankCell{2}{3}{0.082} & \RankCell{2}{3}{0.038} & \RankCell{2}{3}{0.050} & \RankCell{2}{3}{0.042} & \AvgRankCell{2.08}{1.33}{2.08} \\
			$\mathcal{L}_\text{reg}$ + $\mathcal{L}_\text{conf}$ & \RankCell{3}{3}{0.050} & \RankCell{1}{3}{0.064} & \RankCell{2}{3}{0.065} & \RankCell{1}{3}{0.043} & \RankCell{1}{3}{0.046} & \RankCell{1}{3}{0.046} & \RankCell{3}{3}{0.270} & \RankCell{3}{3}{1.059} & \RankCell{2}{3}{0.082} & \RankCell{2}{3}{0.038} & \RankCell{2}{3}{0.050} & \RankCell{3}{3}{0.043} & \AvgRankCell{2.00}{1.33}{2.08} \\
			$\mathcal{L}_\text{reg}(\weight_\text{inv})$ & \RankCell{1}{3}{0.043} & \RankCell{3}{3}{0.066} & \RankCell{1}{3}{0.062} & \RankCell{2}{3}{0.044} & \RankCell{2}{3}{0.050} & \RankCell{1}{3}{0.046} & \RankCell{1}{3}{0.254} & \RankCell{1}{3}{0.866} & \RankCell{1}{3}{0.079} & \RankCell{1}{3}{0.036} & \RankCell{1}{3}{0.039} & \RankCell{1}{3}{0.039} & \AvgRankCell{1.33}{1.33}{2.08} \\
			\midrule
			$\mathcal{L}_\text{reg}(\weight_\text{inv})$ & \RankCell{2}{6}{0.043} & \RankCell{3}{6}{0.066} & \RankCell{4}{6}{0.062} & \RankCell{2}{6}{0.044} & \RankCell{3}{6}{0.050} & \RankCell{4}{6}{0.046} & \RankCell{3}{6}{0.254} & \RankCell{1}{6}{0.866} & \RankCell{3}{6}{0.079} & \RankCell{1}{6}{0.036} & \RankCell{2}{6}{0.039} & \RankCell{1}{6}{0.039} & \AvgRankCell{2.42}{1.92}{5.17} \\
			$\mathcal{L}_\text{reg}(\weight_\text{inv})$ + $\mathcal{L}_\text{sg}$ & \RankCell{5}{6}{0.045} & \RankCell{3}{6}{0.066} & \RankCell{5}{6}{0.063} & \RankCell{4}{6}{0.045} & \RankCell{2}{6}{0.048} & \RankCell{5}{6}{0.048} & \RankCell{6}{6}{0.270} & \RankCell{4}{6}{0.949} & \RankCell{6}{6}{0.082} & \RankCell{5}{6}{0.038} & \RankCell{2}{6}{0.039} & \RankCell{3}{6}{0.041} & \AvgRankCell{4.17}{1.92}{5.17} \\
			$\mathcal{L}_\text{reg}(\weight_\text{inv})$ + $\mathcal{L}_\text{tg}$ & \RankCell{5}{6}{0.045} & \RankCell{5}{6}{0.067} & \RankCell{5}{6}{0.063} & \RankCell{5}{6}{0.046} & \RankCell{3}{6}{0.050} & \RankCell{5}{6}{0.048} & \RankCell{5}{6}{0.263} & \RankCell{6}{6}{1.270} & \RankCell{5}{6}{0.081} & \RankCell{6}{6}{0.039} & \RankCell{6}{6}{0.079} & \RankCell{6}{6}{0.042} & \AvgRankCell{5.17}{1.92}{5.17} \\
			$\mathcal{L}_\text{reg}(\weight_\text{inv})$ + $\mathcal{L}_\text{F}$ & \RankCell{1}{6}{0.042} & \RankCell{1}{6}{0.065} & \RankCell{1}{6}{0.061} & \RankCell{2}{6}{0.044} & \RankCell{3}{6}{0.050} & \RankCell{2}{6}{0.045} & \RankCell{1}{6}{0.245} & \RankCell{5}{6}{1.042} & \RankCell{2}{6}{0.078} & \RankCell{2}{6}{0.037} & \RankCell{5}{6}{0.050} & \RankCell{3}{6}{0.041} & \AvgRankCell{2.33}{1.92}{5.17} \\
			$\mathcal{L}_\text{reg}(\weight_\text{inv})$ + $\mathcal{L}_\text{F}$ + $\mathcal{L}_\text{S}$ & \RankCell{2}{6}{0.043} & \RankCell{6}{6}{0.068} & \RankCell{1}{6}{0.061} & \RankCell{6}{6}{0.047} & \RankCell{6}{6}{0.055} & \RankCell{1}{6}{0.044} & \RankCell{4}{6}{0.255} & \RankCell{2}{6}{0.901} & \RankCell{4}{6}{0.080} & \RankCell{2}{6}{0.037} & \RankCell{1}{6}{0.036} & \RankCell{2}{6}{0.040} & \AvgRankCell{3.08}{1.92}{5.17} \\
			(Our Loss) $\mathcal{L}_\text{reg}(\weight_\text{inv})$ + $\mathcal{L}_\text{F}$ + $\mathcal{L}_\text{consis}$ & \RankCell{2}{6}{0.043} & \RankCell{1}{6}{0.065} & \RankCell{1}{6}{0.061} & \RankCell{1}{6}{0.042} & \RankCell{1}{6}{0.045} & \RankCell{2}{6}{0.045} & \RankCell{2}{6}{0.249} & \RankCell{3}{6}{0.919} & \RankCell{1}{6}{0.077} & \RankCell{2}{6}{0.037} & \RankCell{4}{6}{0.041} & \RankCell{3}{6}{0.041} & \AvgRankCell{1.92}{1.92}{5.17} \\
			\midrule
			w/o VGGT High Resolution & \RankCell{1}{2}{0.043} & \RankCell{1}{2}{0.065} & \RankCell{1}{2}{0.061} & \RankCell{2}{2}{0.042} & \RankCell{2}{2}{0.045} & \RankCell{1}{2}{0.045} & \RankCell{2}{2}{0.249} & \RankCell{1}{2}{0.919} & \RankCell{1}{2}{0.077} & \RankCell{2}{2}{0.037} & \RankCell{2}{2}{0.041} & \RankCell{1}{2}{0.041} & \AvgRankCell{1.42}{1.33}{1.42} \\
			(Ours) w/ Efficient High Resolution & \RankCell{1}{2}{0.043} & \RankCell{2}{2}{0.068} & \RankCell{1}{2}{0.061} & \RankCell{1}{2}{0.038} & \RankCell{1}{2}{0.042} & \RankCell{2}{2}{0.046} & \RankCell{1}{2}{0.238} & \RankCell{2}{2}{0.964} & \RankCell{2}{2}{0.080} & \RankCell{1}{2}{0.031} & \RankCell{1}{2}{0.035} & \RankCell{1}{2}{0.041} & \AvgRankCell{1.33}{1.33}{1.42} \\
			\midrule
			\gray{w/ VGGT High Resolution} & \RankCell{1}{2}{0.056} & \RankCell{2}{2}{0.081} & \RankCell{2}{2}{0.067} & \RankCell{2}{2}{0.057} & \RankCell{2}{2}{0.036} & \RankCell{1}{2}{0.045} & \RankCell{2}{2}{0.237} & \RankCell{2}{2}{0.355} & \RankCell{1}{2}{0.064} & \RankCell{2}{2}{0.039} & \RankCell{2}{2}{0.060} & \RankCell{2}{2}{0.050} & \AvgRankCell{1.75}{1.25}{1.75} \\
			\gray{(Ours) w/ Efficient High Resolution} & \RankCell{2}{2}{0.058} & \RankCell{1}{2}{0.059} & \RankCell{1}{2}{0.061} & \RankCell{1}{2}{0.037} & \RankCell{1}{2}{0.029} & \RankCell{2}{2}{0.046} & \RankCell{1}{2}{0.235} & \RankCell{1}{2}{0.289} & \RankCell{2}{2}{0.071} & \RankCell{1}{2}{0.034} & \RankCell{1}{2}{0.024} & \RankCell{1}{2}{0.042} & \AvgRankCell{1.25}{1.25}{1.75} \\

			\bottomrule
		\end{tabular}}
	\label{tab: ablation_study}
\end{table*}

\begin{table}[th]
	\vspace{-1em}
	\centering
	\small
	\caption{The training data components for ablation study. From ``Data1'' to ``Data3'', the data volume increases progressively. While ``Data1'' and ``Data2'' contain only high-quality data, ``Data3'' incorporates additional noisy data.}
	\vspace{-1em}
	\resizebox{\linewidth}{!}{
		\begin{tabular}{l | l}
			\toprule

			\textbf{Name} & Training Data Components \\

			\midrule
			\multirow{2}{*}{Data1} & Hypersim \cite{roberts2021hypersim}, ScanNet++ \cite{scannet++}, Virtual KITTI2 \cite{cabon2020virtual}, \\
 & MVS-Synth \cite{huang2018deepmvs}, Spring \cite{mehl2023spring}, UnrealStereo4K \cite{unrealstereo4k} \\
			\midrule
			Data2 & ``Data1'', Tartanair \cite{tartanair2020iros}, Parallel Domain \cite{parallel_domain}, TartanGround \cite{patel2025tartanground} \\
			\midrule
			\multirow{2}{*}{Data3} & ``Data2'', ScanNet \cite{dai2017scannet}, ARKitScenes \cite{dehghan2021arkitscenes}, \\
 & GraspNet \cite{fang2020graspnet}, BlendedMVS \cite{yao2020blendedmvs} \\

			\bottomrule
		\end{tabular}
	}
	\label{tab: ablation_components_data}
\end{table}

\noindent\textbf{Training Objective.} 
Using ``Data3'', we ablate the original loss by removing $\mathcal{L}_\text{sg}$ and $\mathcal{L}_\text{conf}$. As shown in Table \ref{tab: ablation_study}, removing either term improves performance. $\mathcal{L}_\text{sg}$ appears to focus excessively on local region variance, resulting in lower overall accuracy. Regarding ($\mathcal{L}_\text{conf}$), the model can find a shortcut: instead of learning difficult regions, it can reduce the overall loss by decreasing the learnable loss weights of these areas. In contrast, using the inverse of depth values as a fixed weight map \cite{MoGe} provides a natural alternative. This strategy focuses on the relatively close areas, and achieves further performance improvement. (``$\mathcal{L}_\text{reg}$ + $\mathcal{L}_\text{conf}$'' vs. ``$\mathcal{L}_\text{reg}(\weight_\text{inv})$''). Besides $\mathcal{L}_\text{sg}$, we observe that the temporal gradient loss $\mathcal{L}_\text{tg}$ \cite{chen2025video} also negatively impacts performance. (``$\mathcal{L}_\text{reg}(\weight_\text{inv})$'' vs. ``$\mathcal{L}_\text{reg}(\weight_\text{inv})$ + $\mathcal{L}_\text{tg}$ '').
\begin{equation}
	\label{eq: tg_loss}
	\mathcal{L}_\text{tg} (\hat{\anyinput}, \anyinput, \weight) = \E_{t \in \T, p \in \M} \left\| \weight_{t,p} \cdot (\nabla_t \hat{\anyinput}_{t,p} - \nabla_t \anyinput_{t,p})  \right\|,
\end{equation}
where $\nabla_t$ represents the temporal difference operation.

\begin{tcolorbox}[
		colframe=black,
		arc=4pt,
		boxsep=1pt,
	]
	\paragraph{\textbf{\textit{Insight} 2.}}
	For the training objectives, gradient-based loss functions and learnable confidence weight maps unexpectedly lead to performance degradation. Instead, adopting a simple inverse depth weight map demonstrates superior effectiveness.
\end{tcolorbox}

To further explore the gap between multi-frame and single-frame methods, we conduct an additional ablation study on key components of the single-frame method. For the alignment strategy in training loss, rather than applying a global scale to align the entire sequence with the ground truth, MoGe \cite{MoGe} applies separate scale-shift alignment for each frame ($L_\text{F}$) and each sampled local 3D spherical region of the point cloud ($L_\text{S}$).
\begin{equation}
	\begin{gathered}
		\mathcal{L}_\text{F}(\hat{\anyinput}, \anyinput, \weight) = \E_{t \in \T, p \in \M} \left\| \weight_{t,p} \cdot (\mathbf{a}_t \cdot \hat{\anyinput}_{t,p} + \mathbf{B}_t - \anyinput_{t,p} ) \right\| , \\
		\mathcal{L}_\text{S}(\hat{\anyinput}, \anyinput, \weight) = \E_{\mathcal S_\text{j} \in \mathcal S, p \in \mathcal S_\text{j}} \left\| \weight_{j,p} \cdot (\mathbf{a}_{j} \cdot \hat{\anyinput}_{j,p} + \mathbf{B}_j - \anyinput_{j,p} ) \right\| , \\
		{\mathcal S}_j = \{ p \mid \|{\point}_p - {\point}_j\| \leq r_j,p\in {\mathcal M} \},\\
	\end{gathered}
\end{equation}
where $\mathbf{a}_t$ and $\mathbf{B}_t$ are the scale and shift alignment parameters for each frame $t$, and $\mathbf{a}_j$ and $\mathbf{B}_j$ are the parameters for each sampled local 3D spherical region $\mathcal{S}_j$, with 3D region radius $r_j$. All these scale-shift parameters are computed with ROE alignment \cite{MoGe}. As presented in Table~\ref{tab: ablation_study}, we observe that supervising with both per-sequence and per-frame alignment improves performance (``$\mathcal{L}_\text{reg}(\weight_\text{inv})$'' vs. ``$\mathcal{L}_\text{reg}(\weight_\text{inv}) + \mathcal{L}_\text{F}$''), while the local region alignment unexpectedly results in a decrease in performance (``$\mathcal{L}_\text{reg}(\weight_\text{inv}) + \mathcal{L}_\text{F}$'' vs. ``$\mathcal{L}_\text{reg}(\weight_\text{inv}) + \mathcal{L}_\text{F} + \mathcal{L}_\text{S}$'').

Moreover, we observe that the predicted depth map, camera parameters, and point cloud do not consistently align with the geometry projection constraint from 2D to 3D. One potential approach is to filter out the inaccurate areas by assessing the inconsistencies. 
In contrast, we propose integrating this inherent geometry constraint directly into the training framework, rather than treating it as a post-processing step. Specifically, we introduce a consistency loss $\mathcal{L}_\text{consis}$ to enforce the alignment between the estimated point cloud and its unprojected counterpart.
\begin{equation}
	\label{eq: consis_loss}
	\begin{gathered}
		\mathcal{L}_\text{consis} (\hat{\point}, \hat{\depth}, \hat{\rot}, \hat{\trans}, \hat{\fov}) = \E_{p \in \M} \left| \mathbf{\hat{P}}_{\text{unproj}}(p) - \mathbf{\hat{P}}(p) \right|  , \\
		\hat{\point}_{\text{unproj}}(p) = \hat{\rotmatrix} (\hat{\depth}(p) \hat{\intr}^{-1} p) +\hat{\trans}, ~\hat{\rotmatrix} = \mathcal{H}(\hat{\rot}), \\
		\hat{\intr} = \mathrm{Intrinsics}(\hat{f}_x, \hat{f}_y, \hat{c}_x, \hat{c}_y), ~\hat{c}_x = W / 2, \hat{c}_y = H / 2 \\
		\hat{f}_x = \frac{W}{2 \tan\left(\hat{\fov}_x / 2\right)}, ~\hat{f}_y = \frac{H}{2 \tan\left(\hat{\fov}_y /2\right)},
	\end{gathered}
\end{equation}
where $\mathrm{Intrinsics}(\cdot, \cdot, \cdot, \cdot)$ computes $3\times3$ camera intrinsic matrix from the focal length and the optical center, and $\mathcal{H}(\cdot)$ transforms a rotation quaternion to a $3\times3$ camera rotation matrix. As shown in Table \ref{tab: ablation_study}, the comparison between ``$\mathcal{L}_\text{reg}(\weight_\text{inv}) + \mathcal{L}_\text{F}$'' and ``$\mathcal{L}_\text{reg}(\weight_\text{inv}) + \mathcal{L}_\text{F} + \mathcal{L}_\text{consis}$'' demonstrates that enforcing consistency can lead to improved robustness and accuracy.

\begin{table}[t]
	\vspace{-1em}
	\centering
	\small
	\caption{The parameter count (Params.) and frames per second (FPS) of VGGT and CARVE tested on a single NVIDIA H200.
		The number of parameters is reported in millions, and the FPS is measured with sequence length 32, averaged over 100 runs after warm-up.}
	\vspace{-1em}
	\begin{tabular}{lcccccc}

		\toprule
		\textbf{Method} & Params. (M) & Image Resolution & FPS \\
		\midrule
		\multirow{2}{*}{VGGT \cite{VGGT}} & \multirow{2}{*}{1189.01} & 518 $\times$ 518 & 24.85 \\
 & & 1036 $\times$ 1036 & 2.54 \\
		\midrule
		CARVE (Ours) & 1214.21 & 1036 $\times$ 1036 & 15.26 \\
		\bottomrule
	\end{tabular}
	\label{tab: params_and_fps}
\end{table}

\begin{table*}[!t]
	\centering
	\small
	\setlength{\tabcolsep}{4pt}
	\caption{Efficiency metrics for different input frames and resolutions on a single NVIDIA H200. The number of floating point operations is measured in teraFLOPs (TFLOPs), and memory is measured in gibibytes (GiB). ``(H$\times$ W)'' represents the input image resolution.
	}
	\vspace{-1 em}
	\begin{tabular}{lcccccc}
		\toprule
 & \multicolumn{2}{c}{\textbf{VGGT \cite{VGGT} (518 $\times$ 518)}} & \multicolumn{2}{c}{\textbf{VGGT \cite{VGGT} (1036 $\times$ 1036)}} & \multicolumn{2}{c}{\textbf{CARVE (Ours, 1036 $\times$ 1036)}}  \\
		\cmidrule(lr){2-3} \cmidrule(lr){4-5} \cmidrule(lr){6-7}
		\textbf{\# Frames} & TFLOPs & Peak GPU Mem (GiB) & TFLOPs & Peak GPU Mem (GiB) & TFLOPs & Peak GPU Mem (GiB) \\
		\midrule
		8 & 25.57 & 8.81 & 101.99 & 21.80 & 52.97 & 9.08 \\
		16 & 51.14 & 10.99 & 203.98 & 30.49 & 105.93 & 11.40 \\
		32 & 102.28 & 15.36 & 407.97 & 47.89 & 211.87 & 16.05 \\
		64 & 204.56 & 25.61 & 815.93 & 88.81 & 423.73 & 25.71 \\
		128 & 409.13 & 46.75 & OOM & OOM & 847.47 & 46.86 \\
		256 & 818.25 & 89.02 & OOM & OOM & 1694.94 & 89.14 \\
		\bottomrule
	\end{tabular}
	\label{tab: system_performance_metrics}
\end{table*}

\begin{figure*}[t!]
	\centering
	\includegraphics[width=\linewidth]{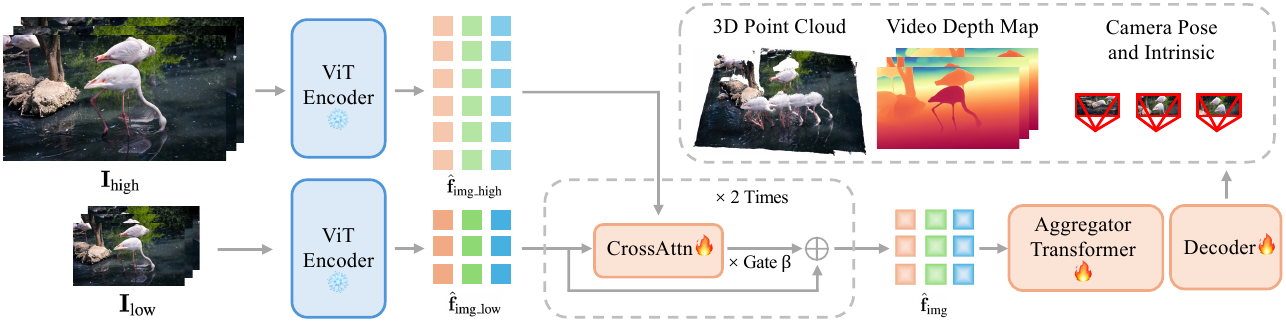}
	\vspace{-1em}
	\caption{Network architecture of our proposed CARVE model. We extract the high-resolution feature and fuse it into the low-resolution main branch with frame-wise cross attention modules and zero-initialized residual gate parameters $\beta$.
	}
	\label{fig: pipeline}
\end{figure*}

\begin{tcolorbox}[
		colframe=black,
		arc=4pt,
		boxsep=1pt,
	]
	\paragraph{\textbf{\textit{Insight} 3.}}
	1) Supervision based on both per-sequence and per-frame alignment enhances the results, while local region alignment unexpectedly leads to a decrease in performance.
	2) Enforcing consistency between the estimated point cloud and the unprojected one yields improvement.

\end{tcolorbox}

\noindent\textbf{Efficient High-Resolution Adaptation.}
It is well recognized that higher-resolution inputs typically enhance the performance of computer vision tasks. However, for the attention module of the transformer block, directly upsampling the input image by a factor of 2 theoretically results in 4$\times$ tokens and 16$\times$ computational complexity. In practice, we report TFLOPs, GPU memory usage, and FPS of VGGT under both low- and high-resolution input settings in Table \ref{tab: system_performance_metrics} and Table \ref{tab: params_and_fps}. Despite the adoption of several engineering optimizations (see the supplementary material for details), high-resolution input still results in 4$\times$ TFLOPs, 3$\times$ to 4$\times$ GPU memory usage, and 0.1$\times$ FPS.

In contrast, we propose an efficient high-resolution adaptation network as illustrated in Figure \ref{fig: pipeline}. We extract the high-resolution feature and fuse it to the low-resolution main branch before sending it to the transformer block with frame-wise cross attention modules.
The low-resolution image serves as the query, while the high-resolution image serves as the key and value. Similar to frame-wise attention, the cross-attention is computed between low- and high-resolution image pairs of the same frame. To prevent the pretrained parameters from being degraded, inspired by ResNet \cite{resnet}, we treat the cross-attention outputs as a residual branch, which is added to the main branch for each cross-attention block after being scaled by a learnable gating parameter. These gating parameters are initialized to zero. For the depth head and point head, we simply upsample the feature prior to the last few convolution layers. The formulation is as follows.
\begin{equation}
	\label{eq: model_ours}
	\begin{gathered}
		\hat{\feat}_{\text{img\_low}} = \mathrm{Encoder}(\image_\text{low}), ~\hat{\feat}_{\text{img\_high}} = \mathrm{Encoder}(\image_\text{high}),  \\
		\hat{\feat}_{\text{img}} = \hat{\feat}_{\text{img\_low}} + \beta \cdot \mathrm{CrossAttn}(\hat{\feat}_{\text{img\_low}}, \hat{\feat}_{\text{img\_high}}), \\
		(\hat{\feat}_{\text{geo}}, \hat{\feat}_{\text{cam}}) = \mathrm{Transformer}(\hat{\feat}_{\text{img}}, \feat_{\text{cam\_init}}), \\
	\end{gathered}
\end{equation}
where the feature $\hat{\feat}_{\text{img\_low}}$ and $\hat{\feat}_{\text{img\_high}}$ are extracted separately from the low-resolution image $\image_\text{low}$ and high-resolution image $\image_\text{high}$, and $\beta$ is the learnable gate parameter with zero initialization. The cross-attention block $\mathrm{CrossAttn}(\cdot, \cdot)$ takes $\hat{\feat}_{\text{img\_low}}$ as query and $\hat{\feat}_{\text{img\_high}}$ as key and value. The fused feature $\hat{\feat}_{\text{img}}$ shares the same dimensionality as $\hat{\feat}_{\text{img\_low}}$, allowing it to seamlessly replace the original low-resolution feature in subsequent modules. As demonstrated in Table \ref{tab: ablation_study}, our architecture enhances the overall performance (``w/o High Resolution'' vs. ``w/ Efficient High Resolution''). Furthermore, our proposed efficient high-resolution architecture even outperforms the direct input upsampling strategy (``w/ Efficient High Resolution'' vs. ``w/ VGGT High Resolution'' in gray color, the evaluations are conducted with a maximum of 100 frames due to GPU memory constraints.). We hypothesize that this improvement arises from two factors: 1) Our efficient architecture processes both high- and low-resolution images, where the integration of multi-resolution features proves beneficial. 2) High-resolution inputs may conflict with the original pretrained weights, which were learned from low-resolution data. For efficiency metrics, our proposed high-resolution architecture achieves substantial computational efficiency, requiring only 0.3$\times$ to 0.4$\times$ GPU memory, 0.5$\times$ TFLOPs, and delivering up to 6$\times$ higher FPS during inference, as reported in Table~\ref{tab: params_and_fps}.
\begin{tcolorbox}[
		colframe=black,
		arc=4pt,
		boxsep=1pt,
	]
	\paragraph{\textbf{\textit{Insight} 4.}}
	Leveraging an efficient high-resolution architecture, the network demonstrates a superior balance between performance and efficiency.

\end{tcolorbox}

\begin{figure*}[t!]
	\centering
	\includegraphics[width=.95\linewidth]{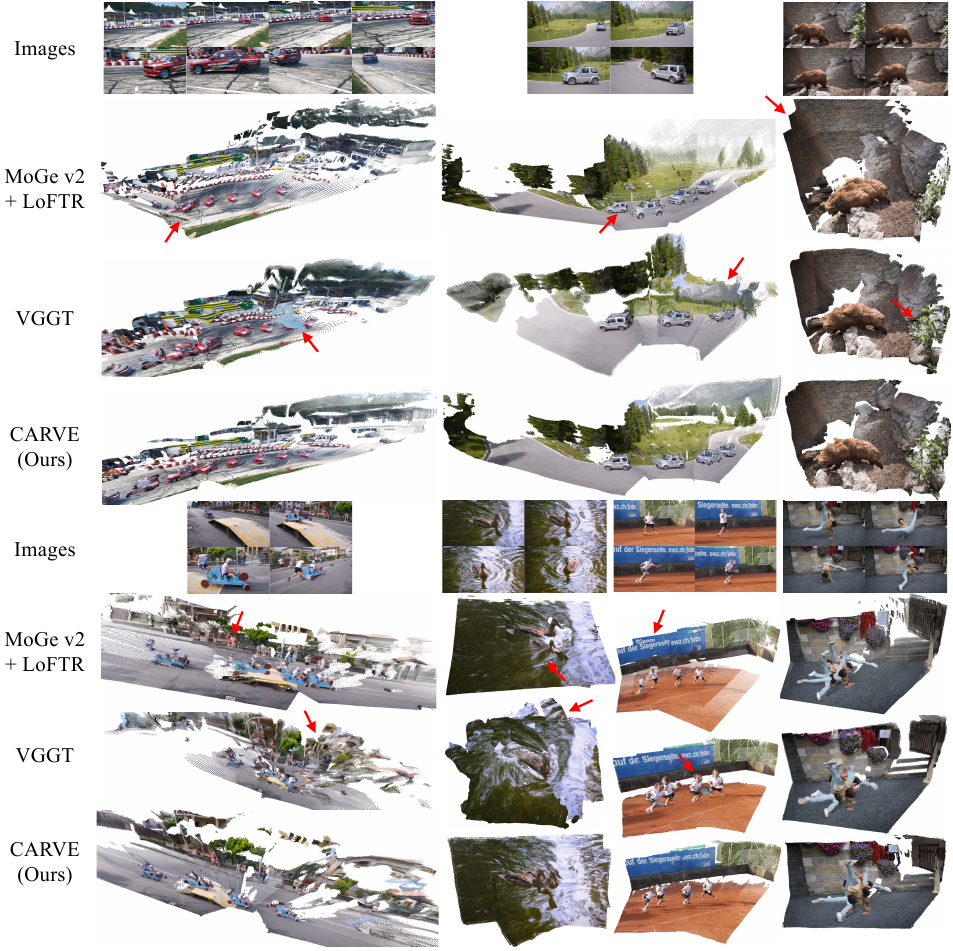}
	\vspace{-1em}
	\caption{Qualitative results of point cloud estimation on in-the-wild images. The red arrows highlight instances of failed estimations, including incorrect camera pose estimation, abnormal geometry scaling, and inconsistencies between frames.
	}
	\label{fig: point_viz}
\end{figure*}

\begin{table*}[!h]
	\centering
	\small
	\setlength{\tabcolsep}{4pt}
	\caption{Quantitative results of point cloud estimation on KITTI, 7-Scenes, and TUM.}
	\vspace{-0.5em}
	\resizebox{\linewidth}{!}{
		\begin{tabular}{lccccccccccccc}
			\toprule
 & \multicolumn{4}{c}{\textbf{KITTI}} & \multicolumn{4}{c}{\textbf{7-Scenes}} & \multicolumn{4}{c}{\textbf{TUM}} & \multirow{2}{*}{\textbf{Rank↓}} \\
			\cmidrule(lr){2-5} \cmidrule(lr){6-9} \cmidrule(lr){10-13}
			\textbf{Method} & C-L1↓ & F@5↑ & F@25↑ & F@50↑ & C-L1↓ & F@5↑ & F@25↑ & F@50↑ & C-L1↓ & F@5↑ & F@25↑ & F@50↑ \\
			\midrule
            MoGe v2 + LoFTR & \RankCell{4}{6}{0.726} & \RankCell{4}{6}{0.142} & \RankCell{4}{6}{0.562} & \RankCell{4}{6}{0.750} & \RankCell{5}{6}{0.161} & \RankCell{5}{6}{0.242} & \RankCell{5}{6}{0.777} & \RankCell{5}{6}{0.950} & \RankCell{5}{6}{0.221} & \RankCell{5}{6}{0.199} & \RankCell{5}{6}{0.696} & \RankCell{5}{6}{0.844} & \AvgRankCell{4.67}{1.42}{5.75} \\
            Spann3R & \RankCell{5}{6}{2.359} & \RankCell{6}{6}{0.044} & \RankCell{6}{6}{0.296} & \RankCell{6}{6}{0.452} & \RankCell{4}{6}{0.101} & \RankCell{4}{6}{0.375} & \RankCell{4}{6}{0.922} & \RankCell{4}{6}{0.987} & \RankCell{4}{6}{0.122} & \RankCell{4}{6}{0.498} & \RankCell{4}{6}{0.860} & \RankCell{4}{6}{0.949} & \AvgRankCell{4.58}{1.42}{5.75} \\
            Fast3R & \RankCell{6}{6}{4.974} & \RankCell{5}{6}{0.088} & \RankCell{5}{6}{0.357} & \RankCell{5}{6}{0.501} & \RankCell{6}{6}{0.655} & \RankCell{6}{6}{0.045} & \RankCell{6}{6}{0.226} & \RankCell{6}{6}{0.422} & \RankCell{6}{6}{0.936} & \RankCell{6}{6}{0.028} & \RankCell{6}{6}{0.153} & \RankCell{6}{6}{0.261} & \AvgRankCell{5.75}{1.42}{5.75} \\
            VGGT & \RankCell{3}{6}{0.296} & \RankCell{3}{6}{0.220} & \RankCell{3}{6}{0.688} & \RankCell{3}{6}{0.842} & \RankCell{2}{6}{0.049} & \RankCell{3}{6}{0.660} & \RankCell{2}{6}{0.988} & \RankCell{2}{6}{0.997} & \RankCell{3}{6}{0.051} & \RankCell{3}{6}{0.712} & \RankCell{3}{6}{0.980} & \RankCell{3}{6}{0.993} & \AvgRankCell{2.75}{1.42}{5.75} \\
            Pi3 & \RankCell{2}{6}{0.273} & \RankCell{1}{6}{0.273} & \RankCell{2}{6}{0.749} & \RankCell{2}{6}{0.879} & \RankCell{2}{6}{0.049} & \RankCell{2}{6}{0.662} & \RankCell{1}{6}{0.991} & \RankCell{2}{6}{0.997} & \RankCell{2}{6}{0.032} & \RankCell{2}{6}{0.834} & \RankCell{1}{6}{0.993} & \RankCell{1}{6}{0.998} & \AvgRankCell{1.67}{1.42}{5.75} \\
            CARVE (Ours) & \RankCell{1}{6}{0.238} & \RankCell{2}{6}{0.257} & \RankCell{1}{6}{0.767} & \RankCell{1}{6}{0.892} & \RankCell{1}{6}{0.043} & \RankCell{1}{6}{0.720} & \RankCell{3}{6}{0.986} & \RankCell{1}{6}{0.998} & \RankCell{1}{6}{0.029} & \RankCell{1}{6}{0.861} & \RankCell{2}{6}{0.991} & \RankCell{2}{6}{0.997} & \AvgRankCell{1.42}{1.42}{5.75} \\
            \bottomrule

		\end{tabular}}
	\label{tab: pointcloud_final_wide_part1}
\end{table*}

\begin{table*}[!t]
	\centering
	\small
	\setlength{\tabcolsep}{4pt}
	\caption{Quantitative results of point cloud estimation on HAMMER, Bonn, and ETH3D.}
	\vspace{-0.5em}
	\resizebox{\linewidth}{!}{
		\begin{tabular}{lccccccccccccc}
			\toprule
 & \multicolumn{4}{c}{\textbf{HAMMER}} & \multicolumn{4}{c}{\textbf{Bonn}} & \multicolumn{4}{c}{\textbf{ETH3D}} & \multirow{2}{*}{\textbf{Rank↓}}
			\\
			\cmidrule(lr){2-5} \cmidrule(lr){6-9} \cmidrule(lr){10-13}
			\textbf{Method} & C-L1↓ & F@5↑ & F@25↑ & F@50↑ & C-L1↓ & F@5↑ & F@25↑ & F@50↑ & C-L1↓ & F@5↑ & F@25↑ & F@50↑
			\\
			\midrule
			MoGe v2 + LoFTR & \RankCell{3}{6}{0.030} & \RankCell{3}{6}{0.872} & \RankCell{1}{6}{1.000} & \RankCell{1}{6}{1.000} & \RankCell{5}{6}{0.174} & \RankCell{5}{6}{0.226} & \RankCell{5}{6}{0.765} & \RankCell{5}{6}{0.941} & \RankCell{4}{6}{1.889} & \RankCell{5}{6}{0.002} & \RankCell{5}{6}{0.026} & \RankCell{5}{6}{0.066} & \AvgRankCell{3.92}{1.17}{5.17} \\
            Spann3R & \RankCell{5}{6}{0.041} & \RankCell{5}{6}{0.727} & \RankCell{1}{6}{1.000} & \RankCell{1}{6}{1.000} & \RankCell{4}{6}{0.114} & \RankCell{4}{6}{0.354} & \RankCell{4}{6}{0.894} & \RankCell{4}{6}{0.978} & \RankCell{5}{6}{2.479} & \RankCell{4}{6}{0.067} & \RankCell{4}{6}{0.216} & \RankCell{4}{6}{0.366} & \AvgRankCell{3.75}{1.17}{5.17} \\
            Fast3R & \RankCell{6}{6}{0.062} & \RankCell{6}{6}{0.488} & \RankCell{1}{6}{1.000} & \RankCell{1}{6}{1.000} & \RankCell{6}{6}{0.983} & \RankCell{6}{6}{0.019} & \RankCell{6}{6}{0.134} & \RankCell{6}{6}{0.285} & \RankCell{6}{6}{3.901} & \RankCell{6}{6}{0.000} & \RankCell{6}{6}{0.000} & \RankCell{6}{6}{0.000} & \AvgRankCell{5.17}{1.17}{5.17} \\
            VGGT & \RankCell{4}{6}{0.035} & \RankCell{4}{6}{0.828} & \RankCell{6}{6}{0.999} & \RankCell{1}{6}{1.000} & \RankCell{3}{6}{0.057} & \RankCell{3}{6}{0.645} & \RankCell{3}{6}{0.972} & \RankCell{3}{6}{0.987} & \RankCell{2}{6}{0.202} & \RankCell{3}{6}{0.410} & \RankCell{2}{6}{0.787} & \RankCell{2}{6}{0.915} & \AvgRankCell{3.00}{1.17}{5.17} \\
            Pi3 & \RankCell{2}{6}{0.013} & \RankCell{2}{6}{0.997} & \RankCell{1}{6}{1.000} & \RankCell{1}{6}{1.000} & \RankCell{1}{6}{0.031} & \RankCell{1}{6}{0.796} & \RankCell{1}{6}{0.998} & \RankCell{1}{6}{1.000} & \RankCell{1}{6}{0.106} & \RankCell{1}{6}{0.433} & \RankCell{1}{6}{0.896} & \RankCell{1}{6}{0.971} & \AvgRankCell{1.17}{1.17}{5.17} \\
            CARVE (Ours) & \RankCell{1}{6}{0.012} & \RankCell{1}{6}{0.999} & \RankCell{1}{6}{1.000} & \RankCell{1}{6}{1.000} & \RankCell{2}{6}{0.043} & \RankCell{2}{6}{0.720} & \RankCell{2}{6}{0.986} & \RankCell{2}{6}{0.998} & \RankCell{3}{6}{0.236} & \RankCell{2}{6}{0.423} & \RankCell{3}{6}{0.765} & \RankCell{3}{6}{0.867} & \AvgRankCell{1.92}{1.17}{5.17} \\

			\bottomrule
		\end{tabular}}
	\label{tab: pointcloud_final_wide_part2}
\end{table*}

\begin{table*}[!t]
	\centering
	\small
	\setlength{\tabcolsep}{4pt}
	\caption{Quantitative results of video depth estimation on seven datasets.}
	\vspace{-0.5em}
	\resizebox{\linewidth}{!}{
		\begin{tabular}{lccccccccccccccc}
			\toprule
 & \multicolumn{2}{c}{\textbf{KITTI}} & \multicolumn{2}{c}{\textbf{7-Scenes}} & \multicolumn{2}{c}{\textbf{TUM}} & \multicolumn{2}{c}{\textbf{HO3D}} & \multicolumn{2}{c}{\textbf{HAMMER}} & \multicolumn{2}{c}{\textbf{Bonn}} & \multicolumn{2}{c}{\textbf{ETH3D}} & \multirow{2}{*}{\textbf{Rank↓}} \\
			\cmidrule(lr){2-3} \cmidrule(lr){4-5} \cmidrule(lr){6-7} \cmidrule(lr){8-9} \cmidrule(lr){10-11} \cmidrule(lr){12-13} \cmidrule(lr){14-15}
			\textbf{Method} & Rel↓ & $\delta$↑ & Rel↓ & $\delta$↑ & Rel↓ & $\delta$↑ & Rel↓ & $\delta$↑ & Rel↓ & $\delta$↑ & Rel↓ & $\delta$↑ & Rel↓ & $\delta$↑ \\
			\midrule
			MoGe v2 + LoFTR & \RankCell{5}{5}{0.453} & \RankCell{5}{5}{0.430} & \RankCell{4}{5}{0.217} & \RankCell{4}{5}{0.675} & \RankCell{4}{5}{0.225} & \RankCell{4}{5}{0.593} & \RankCell{4}{5}{0.278} & \RankCell{3}{5}{0.811} & \RankCell{3}{5}{0.036} & \RankCell{1}{5}{0.997} & \RankCell{4}{5}{0.171} & \RankCell{4}{5}{0.797} & \RankCell{4}{5}{0.242} & \RankCell{4}{5}{0.690} & \AvgRankCell{3.79}{1.50}{4.86} \\
            Fast3R & \RankCell{4}{5}{0.254} & \RankCell{4}{5}{0.645} & \RankCell{5}{5}{0.312} & \RankCell{5}{5}{0.476} & \RankCell{5}{5}{0.377} & \RankCell{5}{5}{0.446} & \RankCell{5}{5}{0.524} & \RankCell{5}{5}{0.587} & \RankCell{5}{5}{0.135} & \RankCell{5}{5}{0.838} & \RankCell{5}{5}{0.339} & \RankCell{5}{5}{0.551} & \RankCell{5}{5}{0.568} & \RankCell{5}{5}{0.335} & \AvgRankCell{4.86}{1.50}{4.86} \\
            VGGT & \RankCell{3}{5}{0.094} & \RankCell{3}{5}{0.917} & \RankCell{3}{5}{0.069} & \RankCell{3}{5}{0.930} & \RankCell{3}{5}{0.062} & \RankCell{3}{5}{0.954} & \RankCell{3}{5}{0.270} & \RankCell{4}{5}{0.755} & \RankCell{4}{5}{0.046} & \RankCell{4}{5}{0.968} & \RankCell{3}{5}{0.054} & \RankCell{3}{5}{0.953} & \RankCell{3}{5}{0.043} & \RankCell{3}{5}{0.978} & \AvgRankCell{3.21}{1.50}{4.86} \\
            Pi3 & \RankCell{1}{5}{0.078} & \RankCell{1}{5}{0.939} & \RankCell{2}{5}{0.064} & \RankCell{2}{5}{0.938} & \RankCell{2}{5}{0.043} & \RankCell{1}{5}{0.977} & \RankCell{2}{5}{0.248} & \RankCell{2}{5}{0.846} & \RankCell{2}{5}{0.033} & \RankCell{3}{5}{0.984} & \RankCell{1}{5}{0.026} & \RankCell{1}{5}{0.987} & \RankCell{1}{5}{0.023} & \RankCell{1}{5}{0.998} & \AvgRankCell{1.57}{1.50}{4.86} \\
            CARVE (Ours) & \RankCell{2}{5}{0.082} & \RankCell{2}{5}{0.933} & \RankCell{1}{5}{0.062} & \RankCell{1}{5}{0.940} & \RankCell{1}{5}{0.040} & \RankCell{2}{5}{0.976} & \RankCell{1}{5}{0.220} & \RankCell{1}{5}{0.869} & \RankCell{1}{5}{0.020} & \RankCell{2}{5}{0.996} & \RankCell{2}{5}{0.041} & \RankCell{2}{5}{0.959} & \RankCell{1}{5}{0.023} & \RankCell{2}{5}{0.997} & \AvgRankCell{1.50}{1.50}{4.86} \\

			\bottomrule
		\end{tabular}}
	\label{tab: depth_perf_final_data_extended}
\end{table*}

\begin{table*}[!t]
	\centering
	\small
	\setlength{\tabcolsep}{1pt}
	\caption{Quantitative evaluation of camera pose and intrinsics on KITTI, 7-Scenes, TUM, and HO3D.}
	\vspace{-0.5em}
	\resizebox{\linewidth}{!}{
		\begin{tabular}{lcccccccccccc c c}
			\toprule
 & \multicolumn{4}{c}{\textbf{KITTI}} & \multicolumn{4}{c}{\textbf{7-Scenes}} & \multicolumn{4}{c}{\textbf{TUM}} & \multicolumn{1}{c}{\textbf{HO3D}} & \multirow{2}{*}{\textbf{Rank↓}}  \\
			\cmidrule(lr){2-5} \cmidrule(lr){6-9} \cmidrule(lr){10-13} \cmidrule(lr){14-14}
			\textbf{Method} & FoV Rel↓ & ATE↓ & RPE-R↓ & RPE-T↓ & FoV Rel↓ & ATE↓ & RPE-R↓ & RPE-T↓ & FoV Rel↓ & ATE↓ & RPE-R↓ & RPE-T↓ & FoV Rel↓ \\
			\midrule
			MoGe v2 & \RankCell{5}{5}{0.162} & -- & -- & -- & \RankCell{5}{5}{0.192} & -- & -- & -- & \RankCell{5}{5}{0.124} & -- & -- & -- & \RankCell{3}{5}{0.067} & \AvgRankCell{4.50}{1.69}{4.50} \\
            Fast3R & \RankCell{2}{5}{0.079} & \RankCell{4}{4}{106.082} & \RankCell{4}{4}{0.161} & \RankCell{4}{4}{125.443} & \RankCell{3}{5}{0.075} & \RankCell{4}{4}{1.696} & \RankCell{4}{4}{1.056} & \RankCell{4}{4}{2.576} & \RankCell{2}{5}{0.028} & \RankCell{4}{4}{1.189} & \RankCell{4}{4}{1.257} & \RankCell{4}{4}{1.897} & \RankCell{1}{5}{0.012} & \AvgRankCell{3.38}{1.69}{4.50} \\
            VGGT & \RankCell{3}{5}{0.084} & \RankCell{3}{4}{1.113} & \RankCell{1}{4}{0.015} & \RankCell{2}{4}{2.177} & \RankCell{4}{5}{0.076} & \RankCell{3}{4}{0.073} & \RankCell{2}{4}{0.062} & \RankCell{3}{4}{0.117} & \RankCell{1}{5}{0.020} & \RankCell{3}{4}{0.047} & \RankCell{3}{4}{0.038} & \RankCell{2}{4}{0.063} & \RankCell{5}{5}{0.109} & \AvgRankCell{2.69}{1.69}{4.50} \\
            Pi3 & \RankCell{4}{5}{0.094} & \RankCell{1}{4}{0.572} & \RankCell{2}{4}{0.016} & \RankCell{3}{4}{2.270} & \RankCell{2}{5}{0.036} & \RankCell{2}{4}{0.058} & \RankCell{1}{4}{0.059} & \RankCell{1}{4}{0.103} & \RankCell{3}{5}{0.045} & \RankCell{2}{4}{0.046} & \RankCell{2}{4}{0.034} & \RankCell{3}{4}{0.071} & \RankCell{4}{5}{0.082} & \AvgRankCell{2.31}{1.69}{4.50} \\
            CARVE (Ours) & \RankCell{1}{5}{0.078} & \RankCell{2}{4}{0.664} & \RankCell{2}{4}{0.016} & \RankCell{1}{4}{1.740} & \RankCell{1}{5}{0.024} & \RankCell{1}{4}{0.052} & \RankCell{3}{4}{0.064} & \RankCell{2}{4}{0.104} & \RankCell{4}{5}{0.049} & \RankCell{1}{4}{0.041} & \RankCell{1}{4}{0.032} & \RankCell{1}{4}{0.060} & \RankCell{2}{5}{0.039} & \AvgRankCell{1.69}{1.69}{4.50} \\

			\bottomrule
		\end{tabular}}
	\label{tab: pose_perf_part1_complete}
\end{table*}

\begin{table*}[!t]
	\centering
	\small
	\setlength{\tabcolsep}{2pt}
	\caption{Quantitative evaluation of camera pose and intrinsics on HAMMER, Bonn, and ETH3D.}
	\vspace{-0.5em}
	\resizebox{\linewidth}{!}{
		\begin{tabular}{lccccccccccccc}
			\toprule
 & \multicolumn{4}{c}{\textbf{HAMMER}} & \multicolumn{4}{c}{\textbf{Bonn}} & \multicolumn{4}{c}{\textbf{ETH3D}} & \multirow{2}{*}{\textbf{Rank↓}} \\
			\cmidrule(lr){2-5} \cmidrule(lr){6-9} \cmidrule(lr){10-13}
			\textbf{Method} & FoV Rel↓ & ATE↓ & RPE-R↓ & RPE-T↓ & FoV Rel↓ & ATE↓ & RPE-R↓ & RPE-T↓ & FoV Rel↓ & ATE↓ & RPE-R↓ & RPE-T↓ \\
			\midrule
			MoGe v2 & \RankCell{5}{5}{0.084} & -- & -- & -- & \RankCell{5}{5}{0.136} & -- & -- & -- & \RankCell{4}{5}{0.058} & -- & -- & -- & \AvgRankCell{4.67}{1.92}{4.67} \\
            Fast3R & \RankCell{3}{5}{0.062} & \RankCell{4}{4}{0.119} & \RankCell{4}{4}{0.166} & \RankCell{4}{4}{0.187} & \RankCell{2}{5}{0.025} & \RankCell{4}{4}{0.669} & \RankCell{4}{4}{0.722} & \RankCell{4}{4}{1.089} & \RankCell{5}{5}{0.075} & \RankCell{4}{4}{13.074} & \RankCell{4}{4}{1.491} & \RankCell{4}{4}{16.466} & \AvgRankCell{3.83}{1.92}{4.67} \\
            VGGT & \RankCell{2}{5}{0.040} & \RankCell{1}{4}{0.001} & \RankCell{1}{4}{0.003} & \RankCell{1}{4}{0.002} & \RankCell{4}{5}{0.040} & \RankCell{3}{4}{0.075} & \RankCell{3}{4}{0.042} & \RankCell{3}{4}{0.091} & \RankCell{2}{5}{0.020} & \RankCell{3}{4}{1.804} & \RankCell{1}{4}{0.021} & \RankCell{3}{4}{2.143} & \AvgRankCell{2.25}{1.92}{4.67} \\
            Pi3 & \RankCell{4}{5}{0.082} & \RankCell{3}{4}{0.003} & \RankCell{3}{4}{0.005} & \RankCell{3}{4}{0.006} & \RankCell{1}{5}{0.022} & \RankCell{1}{4}{0.039} & \RankCell{1}{4}{0.024} & \RankCell{1}{4}{0.055} & \RankCell{3}{5}{0.031} & \RankCell{1}{4}{0.140} & \RankCell{1}{4}{0.021} & \RankCell{1}{4}{0.193} & \AvgRankCell{1.92}{1.92}{4.67} \\
            CARVE (Ours) & \RankCell{1}{5}{0.035} & \RankCell{1}{4}{0.001} & \RankCell{2}{4}{0.004} & \RankCell{2}{4}{0.003} & \RankCell{3}{5}{0.028} & \RankCell{2}{4}{0.044} & \RankCell{2}{4}{0.029} & \RankCell{2}{4}{0.056} & \RankCell{1}{5}{0.018} & \RankCell{2}{4}{0.184} & \RankCell{3}{4}{0.022} & \RankCell{2}{4}{0.223} & \AvgRankCell{1.92}{1.92}{4.67} \\
			\bottomrule
		\end{tabular}}
	\label{tab: pose_perf_part2_complete_extended}
\end{table*}

\begin{table*}[!t]
	\centering
	\small
	\setlength{\tabcolsep}{4pt}
	\caption{Quantitative results of monocular depth estimation on seven datasets.}
	\vspace{-0.5em}
	\resizebox{\linewidth}{!}{
		\begin{tabular}{lccccccccccccccc}
			\toprule
 & \multicolumn{2}{c}{\textbf{KITTI}} & \multicolumn{2}{c}{\textbf{7-Scenes}} & \multicolumn{2}{c}{\textbf{TUM}} & \multicolumn{2}{c}{\textbf{HO3D}} & \multicolumn{2}{c}{\textbf{HAMMER}} & \multicolumn{2}{c}{\textbf{Bonn}} & \multicolumn{2}{c}{\textbf{ETH3D}} & \multirow{2}{*}{\textbf{Rank↓}} \\
			\cmidrule(lr){2-3} \cmidrule(lr){4-5} \cmidrule(lr){6-7} \cmidrule(lr){8-9} \cmidrule(lr){10-11} \cmidrule(lr){12-13} \cmidrule(lr){14-15}
			\textbf{Method} & Rel↓ & $\delta$↑ & Rel↓ & $\delta$↑ & Rel↓ & $\delta$↑ & Rel↓ & $\delta$↑ & Rel↓ & $\delta$↑ & Rel↓ & $\delta$↑ & Rel↓ & $\delta$↑ \\
			\midrule
			MoGe & \RankCell{1}{6}{0.094} & \RankCell{2}{6}{0.904} & \RankCell{3}{6}{0.070} & \RankCell{3}{6}{0.938} & \RankCell{2}{6}{0.055} & \RankCell{2}{6}{0.966} & \RankCell{5}{6}{0.282} & \RankCell{4}{6}{0.788} & \RankCell{2}{6}{0.028} & \RankCell{3}{6}{0.988} & \RankCell{2}{6}{0.034} & \RankCell{1}{6}{0.986} & \RankCell{3}{6}{0.035} & \RankCell{1}{6}{0.988} & \AvgRankCell{2.43}{1.86}{6.00} \\
            MoGe v2 & \RankCell{2}{6}{0.098} & \RankCell{1}{6}{0.908} & \RankCell{5}{6}{0.077} & \RankCell{5}{6}{0.932} & \RankCell{4}{6}{0.057} & \RankCell{3}{6}{0.964} & \RankCell{2}{6}{0.256} & \RankCell{2}{6}{0.837} & \RankCell{1}{6}{0.023} & \RankCell{1}{6}{0.996} & \RankCell{4}{6}{0.037} & \RankCell{3}{6}{0.984} & \RankCell{4}{6}{0.036} & \RankCell{2}{6}{0.986} & \AvgRankCell{2.79}{1.86}{6.00} \\
            Fast3R & \RankCell{6}{6}{0.274} & \RankCell{6}{6}{0.594} & \RankCell{6}{6}{0.247} & \RankCell{6}{6}{0.591} & \RankCell{6}{6}{0.285} & \RankCell{6}{6}{0.597} & \RankCell{6}{6}{0.550} & \RankCell{6}{6}{0.556} & \RankCell{6}{6}{0.143} & \RankCell{6}{6}{0.813} & \RankCell{6}{6}{0.230} & \RankCell{6}{6}{0.646} & \RankCell{6}{6}{0.404} & \RankCell{6}{6}{0.527} & \AvgRankCell{6.00}{1.86}{6.00} \\
            VGGT & \RankCell{5}{6}{0.125} & \RankCell{5}{6}{0.855} & \RankCell{3}{6}{0.070} & \RankCell{4}{6}{0.934} & \RankCell{5}{6}{0.062} & \RankCell{5}{6}{0.948} & \RankCell{3}{6}{0.269} & \RankCell{5}{6}{0.775} & \RankCell{5}{6}{0.054} & \RankCell{5}{6}{0.972} & \RankCell{5}{6}{0.042} & \RankCell{5}{6}{0.975} & \RankCell{5}{6}{0.043} & \RankCell{5}{6}{0.974} & \AvgRankCell{4.64}{1.86}{6.00} \\
            Pi3 & \RankCell{4}{6}{0.112} & \RankCell{4}{6}{0.878} & \RankCell{2}{6}{0.068} & \RankCell{1}{6}{0.941} & \RankCell{2}{6}{0.055} & \RankCell{4}{6}{0.963} & \RankCell{4}{6}{0.271} & \RankCell{3}{6}{0.819} & \RankCell{4}{6}{0.040} & \RankCell{4}{6}{0.986} & \RankCell{1}{6}{0.033} & \RankCell{4}{6}{0.983} & \RankCell{2}{6}{0.034} & \RankCell{2}{6}{0.986} & \AvgRankCell{2.93}{1.86}{6.00} \\
            CARVE (Ours) & \RankCell{3}{6}{0.106} & \RankCell{3}{6}{0.885} & \RankCell{1}{6}{0.066} & \RankCell{1}{6}{0.941} & \RankCell{1}{6}{0.049} & \RankCell{1}{6}{0.969} & \RankCell{1}{6}{0.236} & \RankCell{1}{6}{0.851} & \RankCell{2}{6}{0.028} & \RankCell{2}{6}{0.994} & \RankCell{3}{6}{0.035} & \RankCell{2}{6}{0.985} & \RankCell{1}{6}{0.033} & \RankCell{4}{6}{0.985} & \AvgRankCell{1.86}{1.86}{6.00} \\
			\bottomrule
		\end{tabular}}
	\label{tab: monocular_depth_estimation_results}
\end{table*}

\section{Experiments}
\label{sec: experiments}

We scale up the training process with the losses of ``$\mathcal{L}_\text{reg}(\weight_\text{inv})$ + $\mathcal{L}_\text{F}$ + $\mathcal{L}_\text{consis}$'', training data of ``Data3'', and our proposed efficient high-resolution architecture. The model is initialized with the VGGT \cite{VGGT} pretrained weights for common parameters. More training and evaluation details are provided in the supplementary material. We evaluate visual geometry estimation across multiple datasets, including KITTI \cite{geiger2013vision}, 7-Scenes \cite{shotton2013scene}, HO3D \cite{hampali2020honnotate}, TUM \cite{sturm2012benchmark}, ETH3D \cite{schops2017multi}, HAMMER \cite{jung2023importance}, and Bonn \cite{palazzolo2019iros}.

\subsection{Point Cloud Estimation}

For point cloud evaluation, we align the stacked point cloud with the corresponding stacked ground-truth one using a similarity transformation comprising a scale factor, a rotation matrix, and a translation vector. We report the Chamfer L1 distance (C-L1) and the F-score at thresholds of 5cm (F@5), 25cm (F@25), and 50cm (F@50). The estimated and ground-truth point clouds are downsampled using a voxel size of 2cm for fast evaluation.

We compare with the monocular reconstruction model MoGe v2 \cite{moge2} (``MoGe v2 + LoFTR''), multi-view reconstruction model Spann3R \cite{spann3r}, Fast3R\cite{fast3r}, VGGT \cite{VGGT}, and Pi3 \cite{wang2025pi}. We use LoFTR \cite{sun2021loftr} for feature extraction and matching, and compute the similarity transformation between frames using the matched points to achieve alignment of the results from MoGe v2 \cite{moge2}.

Quantitative comparisons are shown in Table \ref{tab: pointcloud_final_wide_part1} and Table \ref{tab: pointcloud_final_wide_part2}.
As observed, ``MoGe v2 + LoFTR'' performs well on datasets with limited viewpoint variation, such as HAMMER. However, its reliance on accurate matching information limits its effectiveness in more complex scenarios.
VGGT outperforms other multi-view methods, including Spann3R, and Fast3R, which can be attributed to its temporally scalable network framework that improves its ability to model long-range dependencies across views.
Our CARVE, benefiting from our comprehensive analysis and improvements, achieves strong robustness and high accuracy across six evaluation datasets.

Qualitative comparisons are presented in Figure \ref{fig: point_viz}. MoGe v2 + LoFTR demonstrates detailed visualizations but suffers from poor temporal consistency across frames. In contrast, VGGT produces consistent results, but its accuracy is relatively suboptimal. Our proposed method effectively balances both spatial accuracy and temporal consistency, achieving superior overall performance.

\subsection{Video Depth Estimation}

We evaluate video depth using sequence-level scale alignment. Specifically, the predicted depth sequence is aligned to the ground-truth depth sequence with a single global scale factor, and we report the absolute relative error
Rel$=\E_{p \in \M} ~ | (\hat{\depth}_p - \depth_p ) | / \depth_p $
and the percentage of pixels
$\delta=\E_{p \in \M} ~ \max \bigl( \hat{\depth}_p / \depth_p, \depth_p / \hat{\depth}_p \bigr)<1.25$.
Spann3R is not evaluated because it outputs point clouds only in world coordinates.
For MoGe v2, we first align the predicted depth to the sparse LoFTR points on each frame to recover frame-wise depth, and then apply an additional single global scale factor to the whole sequence for fair video-level evaluation.

Quantitative comparisons of video depth estimation are shown in Table \ref{tab: depth_perf_final_data_extended}.
MoGe v2 is limited by inaccurate matching information. Similarly, VGGT outperforms other multi-view methods, including Spann3R and Fast3R. Benefiting from our analysis and improvements, CARVE achieves performance on par with the strongest baseline overall, while outperforming prior methods on several datasets and metrics.

\subsection{Camera Pose and Intrinsics Estimation}

For camera pose estimation, we follow \cite{sturm2012benchmark} to align the predicted camera pose with the ground truth and evaluate the absolute trajectory error (ATE), relative pose error of rotation (RPE-R), and translation (RPE-T). For camera intrinsics, we evaluate the accuracy with the ``FoV Rel'', which is defined as the absolute relative error of the field of view (FoV Rel$=\E_{t} ~ | \hat{\fov}_t - \fov_t | / \fov_t $) to ensure the evaluation of camera intrinsics is independent of image resolution. For MoGe v2, we only evaluate the FoV Rel metric.

Quantitative comparisons are shown in Table \ref{tab: pose_perf_part1_complete} and Table \ref{tab: pose_perf_part2_complete_extended}.
The results show that our proposed CARVE achieves the best overall average rank on KITTI, 7-Scenes, TUM, and HO3D, and ties for the best average rank on HAMMER, Bonn, and ETH3D.

\subsection{Monocular Depth Estimation}

Similar to video depth evaluation, we compare the monocular depth estimation metrics with other feed-forward reconstruction methods.
We continue to use the absolute relative error (Rel.) and the threshold accuracy ($\delta < 1.25$, denoted as $\delta$) for monocular depth estimation, but unlike video depth estimation, we perform per-image alignment.
Quantitative comparisons are shown in Table \ref{tab: monocular_depth_estimation_results}. MoGe exhibits strong performance and demonstrates notable competitiveness in monocular depth estimation tasks. Remarkably, despite not being explicitly optimized for monocular depth estimation, our proposed CARVE achieves competitive performance.

\section{Conclusion}

In this work, we explore the critical factors for visual geometry estimation, focusing on training data, objective design, and high-resolution modeling. Based on these insights, we introduce a consistency loss and a lightweight feature-fusion module for accurate and efficient high-resolution inference. Together, these improvements lead to CARVE, which achieves strong overall performance.

\clearpage 
\section*{Acknowledgments}

{
This work was supported by the National Natural Science Foundation of China (No. 62576315).
}

{
	\small
	\bibliographystyle{ieeenat_fullname}
	\bibliography{main}
}

\clearpage
\setcounter{page}{1}
\maketitlesupplementary

\begin{table*}[t]
\centering
\footnotesize
\setlength{\tabcolsep}{3pt}
\renewcommand{\arraystretch}{0.95}
\caption{Supplementary quantitative comparisons between VGGT finetuned and CARVE (Ours). Colors indicate full-rank colormap within each two-row block.}
\label{tab:supp_vggt_finetuned_vs_carve}
\vspace{-0.6em}

{\footnotesize\textbf{(a) Point cloud estimation on KITTI, 7-Scenes, and TUM.}}
\vspace{0.2em}
\resizebox{\linewidth}{!}{%
\begin{tabular}{lccccccccccccc}
\toprule
& \multicolumn{4}{c}{\textbf{KITTI}} & \multicolumn{4}{c}{\textbf{7-Scenes}} & \multicolumn{4}{c}{\textbf{TUM}} & \multirow{2}{*}{\textbf{Rank}$\downarrow$} \\
\cmidrule(lr){2-5} \cmidrule(lr){6-9} \cmidrule(lr){10-13}
\textbf{Method} & C-L1$\downarrow$ & F@5$\uparrow$ & F@25$\uparrow$ & F@50$\uparrow$
& C-L1$\downarrow$ & F@5$\uparrow$ & F@25$\uparrow$ & F@50$\uparrow$
& C-L1$\downarrow$ & F@5$\uparrow$ & F@25$\uparrow$ & F@50$\uparrow$ & \\
\midrule
VGGT finetuned
& \RankCell{2}{2}{0.250} & \RankCell{1}{2}{0.281} & \RankCell{2}{2}{0.746} & \RankCell{2}{2}{0.886}
& \RankCell{2}{2}{0.048} & \RankCell{2}{2}{0.668} & \RankCell{1}{2}{0.990} & \RankCell{1}{2}{0.998}
& \RankCell{2}{2}{0.034} & \RankCell{2}{2}{0.816} & \RankCell{2}{2}{0.989} & \RankCell{1}{2}{0.997}
& \AvgRankCell{1.67}{1.17}{1.67} \\
CARVE (Ours)
& \RankCell{1}{2}{0.238} & \RankCell{2}{2}{0.257} & \RankCell{1}{2}{0.767} & \RankCell{1}{2}{0.892}
& \RankCell{1}{2}{0.043} & \RankCell{1}{2}{0.720} & \RankCell{2}{2}{0.986} & \RankCell{1}{2}{0.998}
& \RankCell{1}{2}{0.029} & \RankCell{1}{2}{0.861} & \RankCell{1}{2}{0.991} & \RankCell{1}{2}{0.997}
& \AvgRankCell{1.17}{1.17}{1.67} \\
\bottomrule
\end{tabular}}

\vspace{0.45em}
{\footnotesize\textbf{(b) Point cloud estimation on HAMMER, Bonn, and ETH3D.}}
\vspace{0.2em}
\resizebox{\linewidth}{!}{%
\begin{tabular}{lccccccccccccc}
\toprule
& \multicolumn{4}{c}{\textbf{HAMMER}} & \multicolumn{4}{c}{\textbf{Bonn}} & \multicolumn{4}{c}{\textbf{ETH3D}} & \multirow{2}{*}{\textbf{Rank}$\downarrow$} \\
\cmidrule(lr){2-5} \cmidrule(lr){6-9} \cmidrule(lr){10-13}
\textbf{Method} & C-L1$\downarrow$ & F@5$\uparrow$ & F@25$\uparrow$ & F@50$\uparrow$
& C-L1$\downarrow$ & F@5$\uparrow$ & F@25$\uparrow$ & F@50$\uparrow$
& C-L1$\downarrow$ & F@5$\uparrow$ & F@25$\uparrow$ & F@50$\uparrow$ & \\
\midrule
VGGT finetuned
& \RankCell{1}{2}{0.010} & \RankCell{2}{2}{0.995} & \RankCell{1}{2}{1.000} & \RankCell{1}{2}{1.000}
& \RankCell{2}{2}{0.045} & \RankCell{2}{2}{0.684} & \RankCell{1}{2}{0.990} & \RankCell{1}{2}{0.998}
& \RankCell{2}{2}{0.240} & \RankCell{2}{2}{0.383} & \RankCell{2}{2}{0.753} & \RankCell{2}{2}{0.865}
& \AvgRankCell{1.58}{1.17}{1.58} \\
CARVE (Ours)
& \RankCell{2}{2}{0.012} & \RankCell{1}{2}{0.999} & \RankCell{1}{2}{1.000} & \RankCell{1}{2}{1.000}
& \RankCell{1}{2}{0.043} & \RankCell{1}{2}{0.720} & \RankCell{2}{2}{0.986} & \RankCell{1}{2}{0.998}
& \RankCell{1}{2}{0.236} & \RankCell{1}{2}{0.423} & \RankCell{1}{2}{0.765} & \RankCell{1}{2}{0.867}
& \AvgRankCell{1.17}{1.17}{1.58} \\
\bottomrule
\end{tabular}}

\vspace{0.45em}
{\footnotesize\textbf{(c) Video depth estimation on seven datasets.}}
\vspace{0.2em}
\resizebox{\linewidth}{!}{%
\begin{tabular}{lccccccccccccccc}
\toprule
& \multicolumn{2}{c}{\textbf{KITTI}} & \multicolumn{2}{c}{\textbf{7-Scenes}} & \multicolumn{2}{c}{\textbf{TUM}}
& \multicolumn{2}{c}{\textbf{HO3D}} & \multicolumn{2}{c}{\textbf{HAMMER}}
& \multicolumn{2}{c}{\textbf{Bonn}} & \multicolumn{2}{c}{\textbf{ETH3D}} & \multirow{2}{*}{\textbf{Rank}$\downarrow$} \\
\cmidrule(lr){2-3} \cmidrule(lr){4-5} \cmidrule(lr){6-7}
\cmidrule(lr){8-9} \cmidrule(lr){10-11} \cmidrule(lr){12-13} \cmidrule(lr){14-15}
\textbf{Method} & Rel$\downarrow$ & $\delta\uparrow$
& Rel$\downarrow$ & $\delta\uparrow$
& Rel$\downarrow$ & $\delta\uparrow$
& Rel$\downarrow$ & $\delta\uparrow$
& Rel$\downarrow$ & $\delta\uparrow$
& Rel$\downarrow$ & $\delta\uparrow$
& Rel$\downarrow$ & $\delta\uparrow$ & \\
\midrule
VGGT finetuned
& \RankCell{1}{2}{0.082} & \RankCell{1}{2}{0.934}
& \RankCell{2}{2}{0.065} & \RankCell{2}{2}{0.938}
& \RankCell{1}{2}{0.039} & \RankCell{1}{2}{0.977}
& \RankCell{2}{2}{0.248} & \RankCell{2}{2}{0.845}
& \RankCell{2}{2}{0.021} & \RankCell{1}{2}{0.996}
& \RankCell{1}{2}{0.041} & \RankCell{2}{2}{0.957}
& \RankCell{1}{2}{0.023} & \RankCell{2}{2}{0.995}
& \AvgRankCell{1.50}{1.21}{1.50} \\
CARVE (Ours)
& \RankCell{1}{2}{0.082} & \RankCell{2}{2}{0.933}
& \RankCell{1}{2}{0.062} & \RankCell{1}{2}{0.940}
& \RankCell{2}{2}{0.040} & \RankCell{2}{2}{0.976}
& \RankCell{1}{2}{0.220} & \RankCell{1}{2}{0.869}
& \RankCell{1}{2}{0.020} & \RankCell{1}{2}{0.996}
& \RankCell{1}{2}{0.041} & \RankCell{1}{2}{0.959}
& \RankCell{1}{2}{0.023} & \RankCell{1}{2}{0.997}
& \AvgRankCell{1.21}{1.21}{1.50} \\
\bottomrule
\end{tabular}}

\vspace{0.45em}
{\footnotesize\textbf{(d) camera pose and intrinsics on KITTI, 7-Scenes, TUM, and HO3D.}}
\vspace{0.2em}
\resizebox{\linewidth}{!}{%
\begin{tabular}{lcccccccccccccc}
\toprule
& \multicolumn{4}{c}{\textbf{KITTI}} & \multicolumn{4}{c}{\textbf{7-Scenes}} & \multicolumn{4}{c}{\textbf{TUM}} & \multicolumn{1}{c}{\textbf{HO3D}} & \multirow{2}{*}{\textbf{Rank}$\downarrow$} \\
\cmidrule(lr){2-5} \cmidrule(lr){6-9} \cmidrule(lr){10-13} \cmidrule(lr){14-14}
\textbf{Method} & FoV Rel$\downarrow$ & ATE$\downarrow$ & RPE-R$\downarrow$ & RPE-T$\downarrow$
& FoV Rel$\downarrow$ & ATE$\downarrow$ & RPE-R$\downarrow$ & RPE-T$\downarrow$
& FoV Rel$\downarrow$ & ATE$\downarrow$ & RPE-R$\downarrow$ & RPE-T$\downarrow$
& FoV Rel$\downarrow$ & \\
\midrule
VGGT finetuned
& \RankCell{1}{2}{0.074} & \RankCell{2}{2}{1.282} & \RankCell{1}{2}{0.015} & \RankCell{2}{2}{2.369}
& \RankCell{2}{2}{0.036} & \RankCell{2}{2}{0.057} & \RankCell{1}{2}{0.062} & \RankCell{2}{2}{0.105}
& \RankCell{1}{2}{0.043} & \RankCell{2}{2}{0.046} & \RankCell{2}{2}{0.037} & \RankCell{2}{2}{0.065}
& \RankCell{1}{2}{0.039}
& \AvgRankCell{1.62}{1.31}{1.62} \\
CARVE (Ours)
& \RankCell{2}{2}{0.078} & \RankCell{1}{2}{0.664} & \RankCell{2}{2}{0.016} & \RankCell{1}{2}{1.740}
& \RankCell{1}{2}{0.024} & \RankCell{1}{2}{0.052} & \RankCell{2}{2}{0.064} & \RankCell{1}{2}{0.104}
& \RankCell{2}{2}{0.049} & \RankCell{1}{2}{0.041} & \RankCell{1}{2}{0.032} & \RankCell{1}{2}{0.060}
& \RankCell{1}{2}{0.039}
& \AvgRankCell{1.31}{1.31}{1.62} \\
\bottomrule
\end{tabular}}

\vspace{0.45em}
{\footnotesize\textbf{(e) camera pose and intrinsics on HAMMER, Bonn, and ETH3D.}}
\vspace{0.2em}
\resizebox{\linewidth}{!}{%
\begin{tabular}{lcccccccccccccc}
\toprule
& \multicolumn{4}{c}{\textbf{HAMMER}} & \multicolumn{4}{c}{\textbf{Bonn}} & \multicolumn{4}{c}{\textbf{ETH3D}} & \multirow{2}{*}{\textbf{Rank}$\downarrow$} \\
\cmidrule(lr){2-5} \cmidrule(lr){6-9} \cmidrule(lr){10-13}
\textbf{Method} & FoV Rel$\downarrow$ & ATE$\downarrow$ & RPE-R$\downarrow$ & RPE-T$\downarrow$
& FoV Rel$\downarrow$ & ATE$\downarrow$ & RPE-R$\downarrow$ & RPE-T$\downarrow$
& FoV Rel$\downarrow$ & ATE$\downarrow$ & RPE-R$\downarrow$ & RPE-T$\downarrow$ & \\
\midrule
VGGT finetuned
& \RankCell{2}{2}{0.036} & \RankCell{2}{2}{0.002} & \RankCell{1}{2}{0.004} & \RankCell{1}{2}{0.003}
& \RankCell{1}{2}{0.023} & \RankCell{1}{2}{0.038} & \RankCell{1}{2}{0.025} & \RankCell{1}{2}{0.048}
& \RankCell{2}{2}{0.030} & \RankCell{2}{2}{0.233} & \RankCell{2}{2}{0.025} & \RankCell{2}{2}{0.270}
& \AvgRankCell{1.50}{1.33}{1.50} \\
CARVE (Ours)
& \RankCell{1}{2}{0.035} & \RankCell{1}{2}{0.001} & \RankCell{1}{2}{0.004} & \RankCell{1}{2}{0.003}
& \RankCell{2}{2}{0.028} & \RankCell{2}{2}{0.044} & \RankCell{2}{2}{0.029} & \RankCell{2}{2}{0.056}
& \RankCell{1}{2}{0.018} & \RankCell{1}{2}{0.184} & \RankCell{1}{2}{0.022} & \RankCell{1}{2}{0.223}
& \AvgRankCell{1.33}{1.33}{1.50} \\
\bottomrule
\end{tabular}}

\vspace{0.45em}
{\footnotesize\textbf{(f) Monocular depth estimation on seven datasets.}}
\vspace{0.2em}
\resizebox{\linewidth}{!}{%
\begin{tabular}{lccccccccccccccc}
\toprule
& \multicolumn{2}{c}{\textbf{KITTI}} & \multicolumn{2}{c}{\textbf{7-Scenes}} & \multicolumn{2}{c}{\textbf{TUM}} & \multicolumn{2}{c}{\textbf{HO3D}}
& \multicolumn{2}{c}{\textbf{HAMMER}} & \multicolumn{2}{c}{\textbf{Bonn}} & \multicolumn{2}{c}{\textbf{ETH3D}} & \multirow{2}{*}{\textbf{Rank}$\downarrow$} \\
\cmidrule(lr){2-3} \cmidrule(lr){4-5} \cmidrule(lr){6-7} \cmidrule(lr){8-9}
\cmidrule(lr){10-11} \cmidrule(lr){12-13} \cmidrule(lr){14-15}
\textbf{Method} & Rel$\downarrow$ & $\delta\uparrow$
& Rel$\downarrow$ & $\delta\uparrow$
& Rel$\downarrow$ & $\delta\uparrow$
& Rel$\downarrow$ & $\delta\uparrow$
& Rel$\downarrow$ & $\delta\uparrow$
& Rel$\downarrow$ & $\delta\uparrow$
& Rel$\downarrow$ & $\delta\uparrow$ & \\
\midrule
VGGT finetuned
& \RankCell{2}{2}{0.106} & \RankCell{1}{2}{0.889}
& \RankCell{2}{2}{0.070} & \RankCell{2}{2}{0.939}
& \RankCell{2}{2}{0.052} & \RankCell{2}{2}{0.968}
& \RankCell{2}{2}{0.250} & \RankCell{2}{2}{0.833}
& \RankCell{2}{2}{0.029} & \RankCell{1}{2}{0.996}
& \RankCell{2}{2}{0.048} & \RankCell{2}{2}{0.979}
& \RankCell{2}{2}{0.037} & \RankCell{2}{2}{0.981}
& \AvgRankCell{1.86}{1.14}{1.86} \\
CARVE (Ours)
& \RankCell{1}{2}{0.106} & \RankCell{2}{2}{0.885}
& \RankCell{1}{2}{0.066} & \RankCell{1}{2}{0.941}
& \RankCell{1}{2}{0.049} & \RankCell{1}{2}{0.969}
& \RankCell{1}{2}{0.236} & \RankCell{1}{2}{0.851}
& \RankCell{1}{2}{0.028} & \RankCell{2}{2}{0.994}
& \RankCell{1}{2}{0.035} & \RankCell{1}{2}{0.985}
& \RankCell{1}{2}{0.033} & \RankCell{1}{2}{0.985}
& \AvgRankCell{1.14}{1.14}{1.86} \\
\bottomrule
\end{tabular}}

\vspace{-0.4em}
\end{table*}

\section{More Analysis}

\noindent\textbf{Comparison with VGGT Fine-tuned on the Same Data.}
For a fair comparison, we further fine-tune VGGT on ``Data3'' under the same final training setting as CARVE. The model is initialized from the official VGGT pretrained weights and fully fine-tuned for 30K iterations. Other settings, including the optimizer, data preprocessing, and evaluation protocol, follow Sec.~\ref{appendix: experimental_setting_details}. Quantitative results are reported in Table \ref{tab:supp_vggt_finetuned_vs_carve}.

Scaling up the training data significantly improves VGGT. However, CARVE still achieves better overall performance across point cloud estimation, video depth estimation, and camera pose/intrinsic estimation. This suggests that the gains of CARVE come not only from stronger training data, but also from our improved training objective and architecture.

\begin{figure}[t]
\centering    
\includegraphics[width=.93\linewidth]{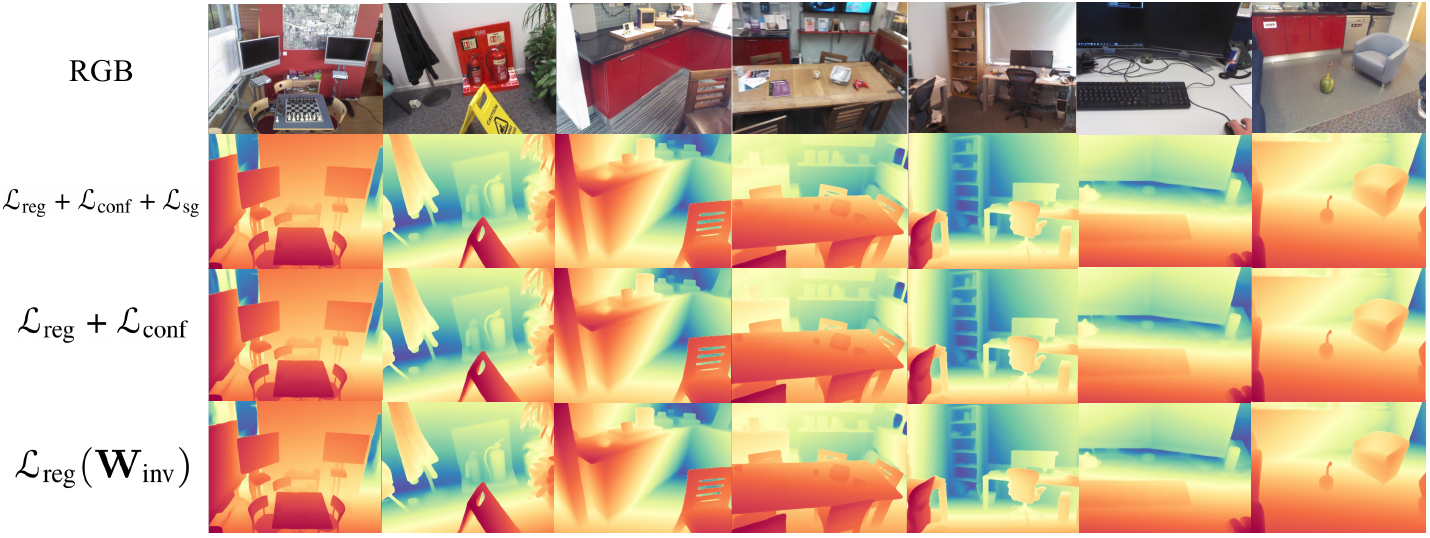}
\caption{Removing the spatial gradient loss and confidence loss has minimal impact on qualitative results when continuing training from VGGT pretrained weights.} 
\label{fig: vis_compare}
\end{figure}

\noindent\textbf{Qualitative Effect of Removing $\mathcal{L}_\text{sg}$ and $\mathcal{L}_\text{conf}$.}
We directly run inference using the models trained with the corresponding loss ablation settings in the main paper. Representative examples are shown in Figure \ref{fig: vis_compare}. Removing $\mathcal{L}_\text{sg}$ and $\mathcal{L}_\text{conf}$ leads to only limited visual differences, while the overall scene geometry and depth structure remain similar. This is consistent with the quantitative results in Table \ref{tab: ablation_study}, suggesting that these terms mainly affect optimization behavior rather than the overall prediction structure.

\begin{table}[t]
\newcommand{\tabincell}[2]{\begin{tabular}{@{}#1@{}}#2\end{tabular}}
  \centering
\small 
\caption{The data type, quality, sequence count and image count of training datasets.}
\resizebox{\linewidth}{!}{
  \begin{tabular}{@{}l|c c c c @{}}
    \toprule

    Training Dataset & Data Type & Data Quality & Sequences & Images \\

    \hline
    ScanNet++ \cite{scannet++} & Indoor & High & 280 & 175661 \\
    Hypersim \cite{roberts2021hypersim} & Indoor & High & 743 & 72019  \\
    ScanNet \cite{dai2017scannet} & Indoor & Middle & 1513 & 2477378 \\
    ARKitScenes \cite{dehghan2021arkitscenes} & Indoor & Middle & 2312 & 2049625 \\
    GraspNet \cite{fang2020graspnet} & Indoor & Middle & 380 & 97280 \\
    Virtual KITTI2 \cite{cabon2020virtual} & Driving & High & 100 & 42520 \\
    MVS-Synth \cite{huang2018deepmvs} & Driving & High & 120 & 12000 \\
    Parallel Domain \cite{parallel_domain} & Driving & High & 367 & 347480 \\
    Spring \cite{mehl2023spring} & Other & High & 74 & 10000 \\
    UnrealStereo4K \cite{unrealstereo4k} & Other & High & 18 & 16400 \\
    Tartanair \cite{tartanair2020iros} & Other & High & 738 & 613274 \\
    TartanGround \cite{patel2025tartanground} & Other & High & 14 & 18484 \\
    BlendedMVS \cite{yao2020blendedmvs} & Other & Middle & 615 & 132961 \\
    \hline
    Total & -- & -- & 7274 & 6065082 \\

    \bottomrule
  \end{tabular}}
  \label{tab: train_dataset_list}
\end{table}

\section{Experimental Setting Details}
\label{appendix: experimental_setting_details}

\begin{figure*}[t]
\centering    
\includegraphics[width=.93\linewidth]{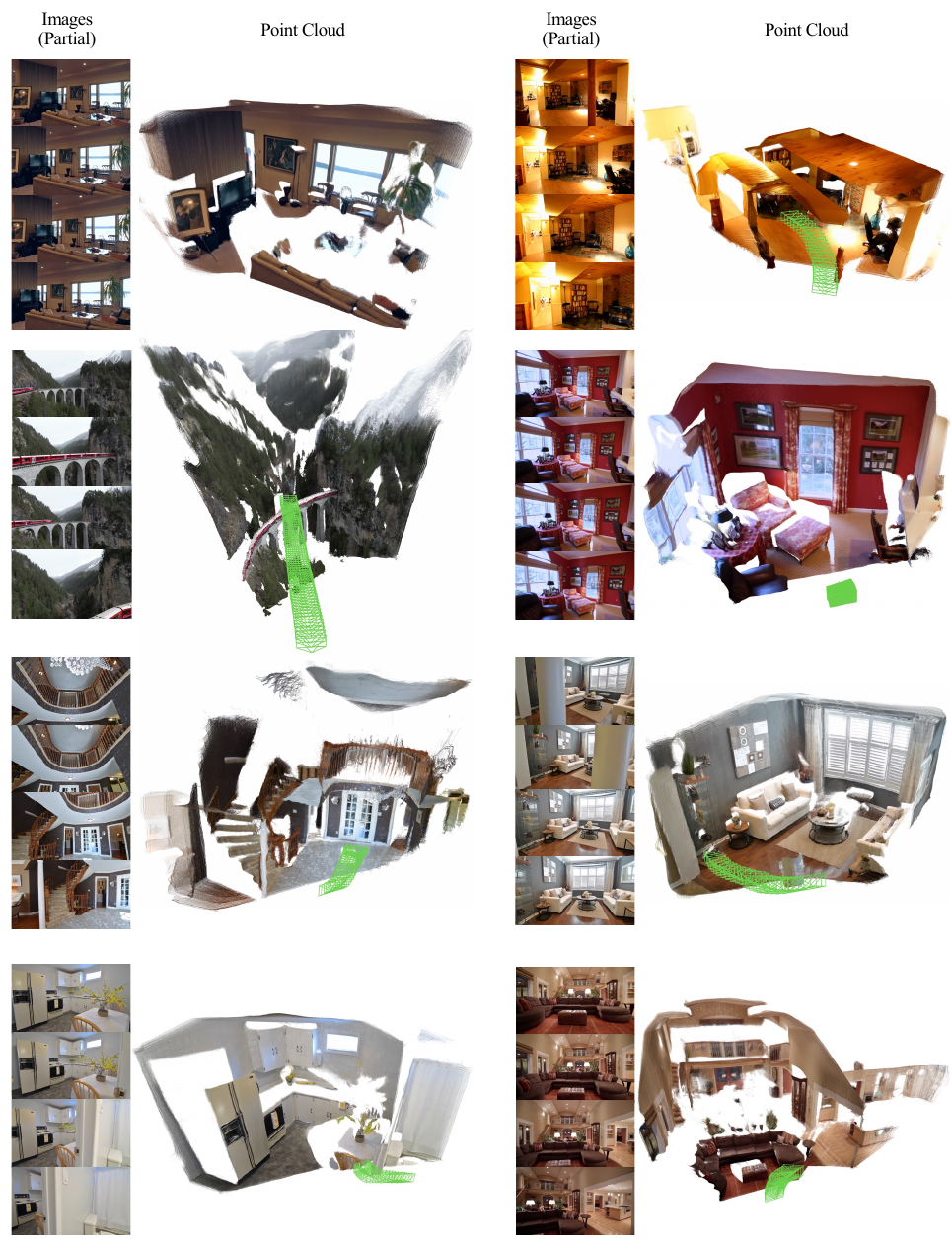}
\caption{More quantitative results of our CARVE model.
} 
\label{fig: point_viz_supp}
\end{figure*}

In the supplementary material, we provide additional details and quantitative results. 1) We present more training and evaluation details for the ablation study and main experiments; 2) We include extended visualization results in Figure \ref{fig: point_viz_supp}.

\noindent\textbf{Common Training Details.}
The experiments were conducted on a server running Ubuntu 22.04 equipped with two Intel Xeon Platinum 8558 CPUs (192 threads in total) and 1.8 TB system memory. The system was configured with eight NVIDIA H200 GPUs using NVIDIA driver 570.133.20 and CUDA 12.4. 

The model is initialized with the VGGT \cite{VGGT} pretrained weights for common parameters. Unless otherwise specified, we freeze the ViT feature extractor and train the remaining components. The regression loss of the camera head is scaled by a factor of 5 to balance between tasks.Training is performed using the AdamW optimizer \cite{loshchilovdecoupled} with $\beta_1 = 0.9$, $\beta_2 = 0.99$, and a weight decay of 0.01. The learning rate is scheduled using the OneCycleLR policy \cite{smith2019super}. The longer side of the low-resolution input image is resized to 518 pixels, and the shorter side is then randomly cropped to one of (448, 378, 308, 238) pixels. For data augmentation, we use random Gaussian blur, Gaussian noise, color jittering, and grayscale. The predicted point cloud, depth map, and camera translation are aligned with the ground-truth values via a scale factor for each sequence before computing the training loss. For the training datasets, they are categorized into three groups: indoor scenes, autonomous driving, and others. To ensure balanced dataset components, we normalize the dataset sizes such that each group contributes an equal volume of data, and individual datasets in each group are expanded to maintain intra-group balance. To accelerate training and reduce CUDA memory requirement, we employ Flash Attention v2 \cite{dao2023flashattention2} in all attention blocks and utilize ZeRO Stage 2 optimization provided by the HuggingFace Accelerate framework. We use PyTorch's gradient checkpointing technique to reduce the CUDA memory usage. A random seed of 2025 is used in our experiment.

\begin{table}[t]
\newcommand{\tabincell}[2]{\begin{tabular}{@{}#1@{}}#2\end{tabular}}
  \centering
\small 
\caption{The data type, sequence count, average frames per sequence of evaluation datasets. ``Stride'' means that we sample every ``Stride'' element from the sequence.}
\resizebox{\linewidth}{!}{
  \begin{tabular}{@{}l|c c c c @{}}
    \toprule

    Eval Dataset & Data Type & Sequences & Avg. Frames & Stride \\

    \hline
    KITTI \cite{geiger2013vision} & Driving & 6 & 107.8 & 1  \\
    7-Scenes \cite{shotton2013scene} & Indoor & 46 & 187.0 & 5  \\
    HO3D \cite{hampali2020honnotate} & Indoor \& Object & 13 & 198.5 & 5 \\
    TUM \cite{sturm2012benchmark} & Indoor & 9 & 199 & 3 \\
    ETH3D \cite{schops2017multi} & Indoor \& Outdoor & 11 & 34.5 & 1 \\
    HAMMER \cite{jung2023importance} & Indoor \& Object & 9 & 130.0 & 1 \\
    Bonn \cite{palazzolo2019iros} & Indoor & 26 & 185.7 & 3 \\
    \bottomrule
  \end{tabular}}
  \label{tab: eval_dataset_list}
\end{table}

\noindent\textbf{Training Details for Ablation Study.}
The model is trained with a learning rate of 3e-6 for 30K iterations on a single NVIDIA H200 GPU, and we employ a dynamic batch size with the sequence length varying between 2 and 24 frames. We constrain the total number of input images to a maximum of 24 for each iteration.

For the training data ablation study, we adopt the original loss component $\mathcal{L}_\text{reg}$ + $\mathcal{L}_\text{sg}$ + $\mathcal{L}_\text{conf}$. For the training objective ablation, we use the training dataset component of ``Data3'', including ScanNet++ \cite{scannet++}, Hypersim \cite{roberts2021hypersim}, ScanNet \cite{dai2017scannet},  ARKitScenes \cite{dehghan2021arkitscenes}, GraspNet \cite{fang2020graspnet}, Virtual KITTI2 \cite{cabon2020virtual}, MVS-Synth \cite{huang2018deepmvs}, Parallel Domain \cite{parallel_domain}, Spring \cite{mehl2023spring}, UnrealStereo4K \cite{unrealstereo4k}, Tartanair \cite{tartanair2020iros}, TartanGround \cite{patel2025tartanground},  and BlendedMVS \cite{yao2020blendedmvs}.

\noindent\textbf{Training Details for CARVE.}
Based on the preceding analysis, we adopt our efficient high-resolution adaptation with two cross-attention blocks to handle the input high-resolution image. 

During training, the corresponding high-resolution image maintains twice the resolution of the low-resolution input. For each GPU, we employ a dynamic batch size with the sequence length varying between 2 and 50 frames, and constrain the total number of input images to a maximum of 50 for each iteration. We use a learning rate of 1e-5, and train the model for 30K iterations on 8 NVIDIA H200 GPUs. For training loss, we adopt the final loss plan of ``$\mathcal{L}_\text{reg}(\weight_\text{inv})$ + $\mathcal{L}_\text{F}$ + $\mathcal{L}_\text{consis}$''.

For training data, we leverage a diverse set of datasets same to ``Data3'', including ScanNet++ \cite{scannet++}, Hypersim \cite{roberts2021hypersim}, ScanNet \cite{dai2017scannet},  ARKitScenes \cite{dehghan2021arkitscenes}, GraspNet \cite{fang2020graspnet}, Virtual KITTI2 \cite{cabon2020virtual}, MVS-Synth \cite{huang2018deepmvs}, Parallel Domain \cite{parallel_domain}, Spring \cite{mehl2023spring}, UnrealStereo4K \cite{unrealstereo4k}, Tartanair \cite{tartanair2020iros}, TartanGround \cite{patel2025tartanground},  and BlendedMVS \cite{yao2020blendedmvs}. The training datasets are listed in Table \ref{tab: train_dataset_list}.

\noindent\textbf{Evaluation Details.} To demonstrate the generalization capability of each method and assess their practical applicability, we evaluate visual geometry estimation across multiple datasets, including KITTI \cite{geiger2013vision}, 7-Scenes \cite{shotton2013scene}, HO3D \cite{hampali2020honnotate}, TUM \cite{sturm2012benchmark}, ETH3D \cite{schops2017multi}, HAMMER \cite{jung2023importance}, and Bonn \cite{palazzolo2019iros}. For the ablation study, we evaluate on 7-Scenes, Bonn, KITTI, and TUM. 
Similar to the evaluation of FrozenRecon \cite{xu2023frozenrecon}, each dataset comprises multiple sequences, from which we uniformly sample keyframes for evaluation using a pre-defined stride between consecutive frames, with a maximum keyframe number of 200. We perform evaluations on the NVIDIA H200 GPU. Due to CUDA memory constraints, we limit the number of frames per sequence to a maximum of 200. The evaluation datasets are listed in Table \ref{tab: eval_dataset_list}. For KITTI, we use the sequences of 2011\_09\_26\_0001, 2011\_09\_26\_0009, 2011\_09\_26\_0091, 2011\_09\_28\_0001, 2011\_09\_29\_0004, and 2011\_09\_29\_0071.

For point cloud estimation, we aggregate the predictions of each sequence in world coordinates by stacking the individual estimations. To assess both per-view accuracy and cross-view consistency, we align the stacked predicted point cloud with the corresponding stacked ground-truth point cloud using a similarity transformation comprising a scale factor, a rotation matrix, and a translation vector. 

For camera pose translation vector and video depth estimation, we align the predictions with the ground truth through a scale value for each sequence. For monocular depth estimation, we align a scale value for each image.

\end{document}